\documentclass[11pt]{article}

\usepackage[final]{acl}

\usepackage{times}
\usepackage{latexsym}

\usepackage[T1]{fontenc}

\usepackage[utf8]{inputenc}

\usepackage{microtype}

\usepackage{inconsolata}

\usepackage{graphicx} 
\usepackage{natbib}  
\usepackage{caption} 
\usepackage{algorithm}
\usepackage{algorithmic}
\usepackage{amsmath}
\usepackage{amssymb}
\usepackage[most]{tcolorbox}
\tcbuselibrary{skins, breakable}   
\usepackage{newfloat}
\usepackage{listings}
\usepackage{xcolor}      
\usepackage{colortbl}    
\usepackage{multirow}    
\usepackage{array}       
\usepackage{booktabs}    
\DeclareCaptionStyle{ruled}{labelfont=normalfont,labelsep=colon,strut=off} 
\floatstyle{ruled}
\newfloat{listing}{tb}{lst}{}
\floatname{listing}{Listing}

\definecolor{metricblue}{RGB}{37,99,235}
\definecolor{metricred}{RGB}{220,38,38}
\definecolor{metricteal}{RGB}{0,128,128}
\definecolor{egcpshade}{RGB}{214,226,238}
\newtcolorbox{definitionbox}[1]{
  enhanced,
  colback=black!3,
  colframe=black,
  colbacktitle=black!12,
  coltitle=black,
  fonttitle=\bfseries,
  title={#1},
  boxrule=0.6pt,
  arc=1mm,
  left=6pt,
  right=6pt,
  top=5pt,
  bottom=5pt,
  before skip=6pt,
  after skip=6pt
}
\newcommand{\ClozeDown}{\shortstack{\textbf{Cloze}\\\textbf{Acc. (\textcolor{metricteal}{$\downarrow$})}}}
\newcommand{\ClassDown}{\shortstack{\textbf{Class.}\\\textbf{Acc. (\textcolor{metricblue}{$\downarrow$})}}}
\newcommand{\RougeDown}{\shortstack{\textbf{Gene. ROUGE-L}\\\textbf{(\textcolor{metricred}{$\downarrow$})}}}
\newcommand{\ClozeUp}{\shortstack{\textbf{Cloze}\\\textbf{Acc. (\textcolor{metricteal}{$\uparrow$})}}}
\newcommand{\ClassUp}{\shortstack{\textbf{Class.}\\\textbf{Acc. (\textcolor{metricblue}{$\uparrow$})}}}
\newcommand{\RougeUp}{\shortstack{\textbf{Gene. ROUGE-L}\\\textbf{(\textcolor{metricred}{$\uparrow$})}}}

\title{Toward Fine-Grained Forgetting:\\ Attribute Unlearning for Multimodal Large Language Models}
\author{
    Junkai Lin\textsuperscript{1,2},
    Junkai Chen\textsuperscript{1} ,
    Siqi Hou\textsuperscript{3,4}, \\
    Yuhao He\textsuperscript{1},
    Ruiqi Liu\textsuperscript{1,5},
    Chenhan Jin\textsuperscript{2}, 
    Shengze Xu\textsuperscript{2},
    Tieyong Zeng\textsuperscript{6},
    Xiangyuan Lan*\textsuperscript{7}
    \\[1em]
    \textsuperscript{1} Institute of Automation, Chinese Academy of Sciences 
    \textsuperscript{2} The Chinese University of Hong Kong \\
    \textsuperscript{3} The Australian National University 
    \textsuperscript{4} Hong Kong Baptist University \\
    \textsuperscript{5} University of Chinese Academy of Sciences 
    \textsuperscript{6} Beijing Normal-Hong Kong Baptist University \\
    \textsuperscript{7} Pengcheng Laboratory, Shenzhen \\
    \texttt{jackielin@link.cuhk.edu.hk}, \texttt{lanxy@pcl.ac.cn}
}

\begin{document}
\maketitle

\begin{abstract}
Multimodal large language models (MLLMs) can memorize and disclose sensitive information, motivating machine unlearning to remove designated knowledge while preserving model utility. Existing privacy-oriented benchmarks mainly study profile-level deletion, whereas practical requests often require finer-grained forgetting: removing a sensitive attribute while retaining other information about the same identity. We introduce attribute-level MLLM unlearning and construct AMU-Bench, covering long-text, numeric, and short-text targets, multiple forget ratios, and diverse question types. Our evaluation shows that target and retain attributes share identity-specific and visual evidence, making selective forgetting prone to residual leakage and collateral degradation. To address this challenge, we propose Causal Localization and Retain-Aware Projection (CLRP), a lightweight, training-free framework that localizes causally relevant layers and projects intermediate representations away from target-specific subspaces while preserving retain-related information. Experiments across MLLMs with different architectures and scales show that CLRP achieves a strong forgetting--retention trade-off on AMU-Bench. For example, at a 5\% forget ratio, CLRP reduces Attribute Forget Set cloze accuracy from 20.00\% to 4.00\% while improving Attribute Retain Set accuracy from 14.77\% to 17.79\% in the reported LLaVA-1.5-7B and Qwen2.5-VL-3B results. These findings demonstrate effective and selective attribute-level unlearning without modifying model parameters.

\end{abstract}

\section{Introduction}

Multimodal large language models (MLLMs), exemplified by LLaVA~\cite{llava} and Qwen2.5-VL~\cite{qwen25vl}, have exhibited strong capabilities in visual question answering~\cite{vqa}, image-grounded instruction following~\cite{llava}, document understanding, and multimodal reasoning~\cite{qwen25vl}. However, large-scale multimodal pretraining may also cause models to memorize and disclose privacy-sensitive identity information~\cite{mllmu,fiubench} or other sensitive knowledge elicited through multimodal queries~\cite{unlokvqa}. Machine unlearning~\cite{llmsurvey,mllmsurvey} seeks to remove designated knowledge from a trained model without retraining it from scratch while preserving general utility. Training-based unlearning methods~\cite{ga,npo,multidelete} update model parameters and may consequently degrade capabilities beyond the deletion target. By contrast, training-free interventions~\cite{kvm,mllmeraser} leave model parameters unchanged but may provide insufficient selectivity when target knowledge is strongly supported by visual evidence.

Representative MLLM unlearning benchmarks define deletion at the granularity of an entity: MLLMU-Bench~\cite{mllmu} uses profiles, FIUBench~\cite{fiubench} uses fictitious identities, and CLEAR~\cite{clear} uses characters. In their standard protocols, the designated entity constitutes the complete forget unit; consequently, evaluation does not explicitly require preservation of individual non-target attributes within that entity. Practical privacy requests, however, are often finer grained. A user may request deletion of a single birth date, address-like field, numeric identifier, organizational affiliation, or biographical description, while benign attributes and general visual recognition of the same identity should remain available. Motivated by this mismatch, we introduce attribute-level MLLM unlearning, which requires the designated attribute to become inaccessible while preserving non-target information associated with the same identity.

Compared with profile-level unlearning, attribute-level unlearning imposes a substantially stricter selectivity requirement. At the data level, conventional forget and retain sets typically contain different profiles, whereas Attribute Forget Set and Attribute Retain Set examples share the same identity and image and often follow closely matched prompt structures. At the representation level, this shared context can place target and non-target attributes in nearby regions of representation space and route them through overlapping causal pathways~\cite{mip-editor,cagul,smfa}. The resulting entanglement makes the boundary between information to remove and information to preserve difficult to identify. An intervention broad enough to eliminate the target attribute may therefore erase neighboring non-target attributes, whereas an overly conservative intervention may leave target information accessible. The central challenge is to isolate and remove the target-specific attribute direction while preserving the shared identity representation required by non-target information.

Building on the profile data of the existing dataset, we introduce the Attribute-level Multimodal Unlearning Benchmark (AMU-Bench), which partitions deletion targets into long-text, numeric, and short-text categories and measures both target forgetting and same-profile preservation. We evaluate three forget ratios, five evaluation views, and three question types on LLaVA-1.5-7B, LLaVA-1.5-13B~\cite{llava} and Qwen2.5-VL-3B~\cite{qwen25vl}. Our results show that existing approaches do not achieve reliable attribute-level selectivity: standard unlearning objectives~\cite{ga,gd,klmin,po} exhibit an unfavorable trade-off between reducing leakage on Attribute Forget Set and preserving utility on Attribute Retain Set. We therefore propose Causal Localization and Retain-Aware Projection (CLRP), a training-free method that combines causal layer selection, task-aligned contrastive features, and retain-aware subspace projection. Averaged over the three attribute categories on LLaVA-1.5-7B and Qwen2.5-VL-3B, CLRP reduces Attribute Forget Set cloze accuracy from 14.50\% to 2.00\% while increasing Attribute Retain Set cloze accuracy from 25.25\% to 29.76\%. Our contributions are summarized below:
\begin{itemize}
    \item We define attribute-level MLLM unlearning as a finer-grained deletion setting in which designated attributes must become inaccessible while non-target facts of the same identity remain available.
    \item We construct AMU-Bench with long-text, numeric, and short-text attribute categories and systematically evaluate representative unlearning methods~\cite{ga,gd,klmin,po} on widely used MLLMs~\cite{llava,qwen25vl}. The evaluation shows that preserving Attribute Retain Set knowledge is more challenging than preserving Shared Retain Set knowledge and that forgetting difficulty varies across attribute categories.
    \item We propose CLRP, a training-free representation intervention that combines patching-based causal layer localization, task-aligned feature construction, and retain-aware contrastive subspace projection. In the 5\% results for LLaVA-1.5-7B and Qwen2.5-VL-3B reported in the main paper, CLRP reduces Attribute Forget Set cloze accuracy by up to 16.00 percentage points and increases Attribute Retain Set cloze accuracy by up to 3.02 percentage points relative to the strongest baseline in each condition.
\end{itemize}

\section{Related Work}

\paragraph{Machine Unlearning in MLLMs.}
MLLM unlearning has expanded across both deletion targets and evaluation settings. Existing work considers single-image removal~\cite{siu}, unsafe or adversarial visual concepts~\cite{safeeraser,auvic}, privacy-sensitive profiles and associations~\cite{mllmu,unlokvqa}, facial identities and characters~\cite{fiubench,clear}, copyrighted knowledge~\cite{covubench}, misinformation~\cite{offside}, and sensitive concept associations~\cite{salmubench}. Recent benchmarks further examine attack--defense robustness~\cite{unlokvqa}, practical deployment scenarios~\cite{pulse}, modality-sensitive evaluation~\cite{umubench}, personalized partial deletion~\cite{ppubench}, continual unlearning~\cite{icubench}, and reasoning preservation~\cite{rmllmubench}. In parallel, method-oriented studies have investigated efficient single-layer updates~\cite{cai2024targeted}, cross-modal safety transfer through textual unlearning~\cite{chakraborty2024cross}, domain-disentangled VLM unlearning~\cite{kawamura2026approximate}, encoder-level reinforcement unlearning~\cite{jia2026object}, and influence-guided cross-modal localization~\cite{balaji2026one}, showing that reliable MLLM unlearning depends not only on the deletion target but also on where and how the intervention is applied. Among profile-oriented benchmarks, MLLMU-Bench~\cite{mllmu}, FIUBench~\cite{fiubench}, and CLEAR~\cite{clear} are most closely related to our data setting because they jointly evaluate target forgetting, retained utility, and held-out recovery. Their standard protocols define the forget unit as a complete profile, identity, or character and primarily assess preservation outside that designated unit; they do not explicitly test whether individual non-target attributes of the selected entity remain accessible. AMU-Bench retains this multi-view evaluation principle but defines each deletion target as a named attribute and introduces Attribute Retain Set to measure preservation within the selected profile. This design isolates same-profile collateral damage that entity-level retention sets may fail to reveal.

\paragraph{Causal tracing and knowledge localization.}
Knowledge editing methods first seek the internal components that causally mediate a target prediction. Causal tracing restores hidden states from a clean run into a corrupted run and measures the resulting recovery of the target output, thereby distinguishing mediating sites from activations that are merely correlated with the prediction~\cite{rome}. Subsequent methods localize influential neurons and cross-layer paths~\cite{mip-editor}, identify target-relevant attention components~\cite{cagul}, or select memory components for localized multimodal forgetting~\cite{smfa}. These approaches improve intervention locality, but entity-level localization can still assign importance to pathways that encode broadly shared identity information. Attribute-level unlearning requires a finer causal contrast because the target attribute and the attributes to preserve may be represented in nearby locations. 

\paragraph{SVD-based subspace projection.}
Linear concept-removal methods identify a low-dimensional target subspace and attenuate its contribution to hidden representations. INLP iteratively removes linearly decodable directions~\cite{inlp}, LEACE derives a closed-form covariance transformation~\cite{leace}, and multimodal unlearning methods estimate forget subspaces from singular-value or covariance structure~\cite{kvm,ccup,dau,sineproject}. Because a forget-only basis can also capture directions supporting nearby non-target attributes, CLRP estimates its projection at a causally selected layer under retain-derived constraints.

\section{Definition}
\label{sec:definition}

AMU-Bench represents each profile as $p=(I_p,\mathcal{A}_p)$, where $I_p$ is the associated image and $\mathcal{A}_p=\{(a_j,v_j)\}_{j=1}^{J_p}$ is a set of named attributes $a_j$ and values $v_j$. For a selected profile $p$, a deletion request specifies a proper subset $\mathcal{A}_p^F\subsetneq\mathcal{A}_p$ and requires its complement $\mathcal{A}_p^R=\mathcal{A}_p\setminus\mathcal{A}_p^F$ to remain accessible. The basic deletion unit is therefore a profile--attribute pair; a request may contain one or more such pairs but does not designate the complete profile for removal.

\begin{definitionbox}{Definition 1 (Attribute-Level MLLM Unlearning)}
\textit{Attribute-Level MLLM Unlearning} removes an MLLM's ability to disclose designated profile attributes, including under held-out reformulations, while preserving the same profiles' non-target attributes and unrelated multimodal knowledge.

\end{definitionbox}

\begin{center}
\scalebox{0.9}{%
\footnotesize
\begin{tabular}{ll}
\toprule
\textbf{Item} & \textbf{Protocol} \\
\midrule
Deletion unit & profile--attribute pair \\
Forget ratio & fraction of selected profiles \\
Operator scope & per attribute category and interface \\
Hook trigger & corresponding evaluation interface \\
Forget queries & target attributes of selected profiles \\
Attr. retain queries & non-target attributes of selected profiles \\
Shared retain queries & profiles outside deletion requests \\
Test queries & held-out target-attribute reformulations \\
\bottomrule
\end{tabular}%
}
\end{center}
This definition imposes both generalization and locality requirements. First, the designated attribute must remain inaccessible under held-out reformulations rather than only under the deletion prompt, excluding prompt-specific blocking as a sufficient solution. Second, because the deletion unit is a profile--attribute pair, the remaining attributes of the selected profile must retain their accessibility; suppressing the complete profile or identity therefore does not satisfy attribute-level unlearning. Preservation of unrelated knowledge further distinguishes a selective intervention from broad utility degradation.

\subsection{Evaluation Set Construction}
For the selected profiles $\mathcal{P}_F$, the \textbf{Attribute Forget Set} $\mathcal{D}_{\mathrm{AF}}$ contains queries whose answers express values in $\mathcal{A}_p^F$; it measures recovery of the designated attributes after unlearning. The \textbf{Attribute Retain Set} $\mathcal{D}_{\mathrm{AR}}$ queries values in $\mathcal{A}_p^R$ for the same profiles and measures preservation of their non-target facts. The held-out \textbf{Test Set} $\mathcal{D}_{\mathrm{Test}}$ contains unseen reformulations targeting $\mathcal{A}_p^F$. The \textbf{Shared Retain Set} $\mathcal{D}_{\mathrm{SR}}$ is drawn from unselected fictitious profiles, and the \textbf{Real-Person Retain Set} $\mathcal{D}_{\mathrm{RP}}$ contains real-person profiles outside the deletion requests. Accordingly, the Attribute Retain Set evaluates within-profile preservation, while the Shared Retain Set and Real-Person Retain Set evaluate preservation outside the selected profiles.

\begin{figure*}[t]
    \centering
    \includegraphics[width=0.82\textwidth]
    {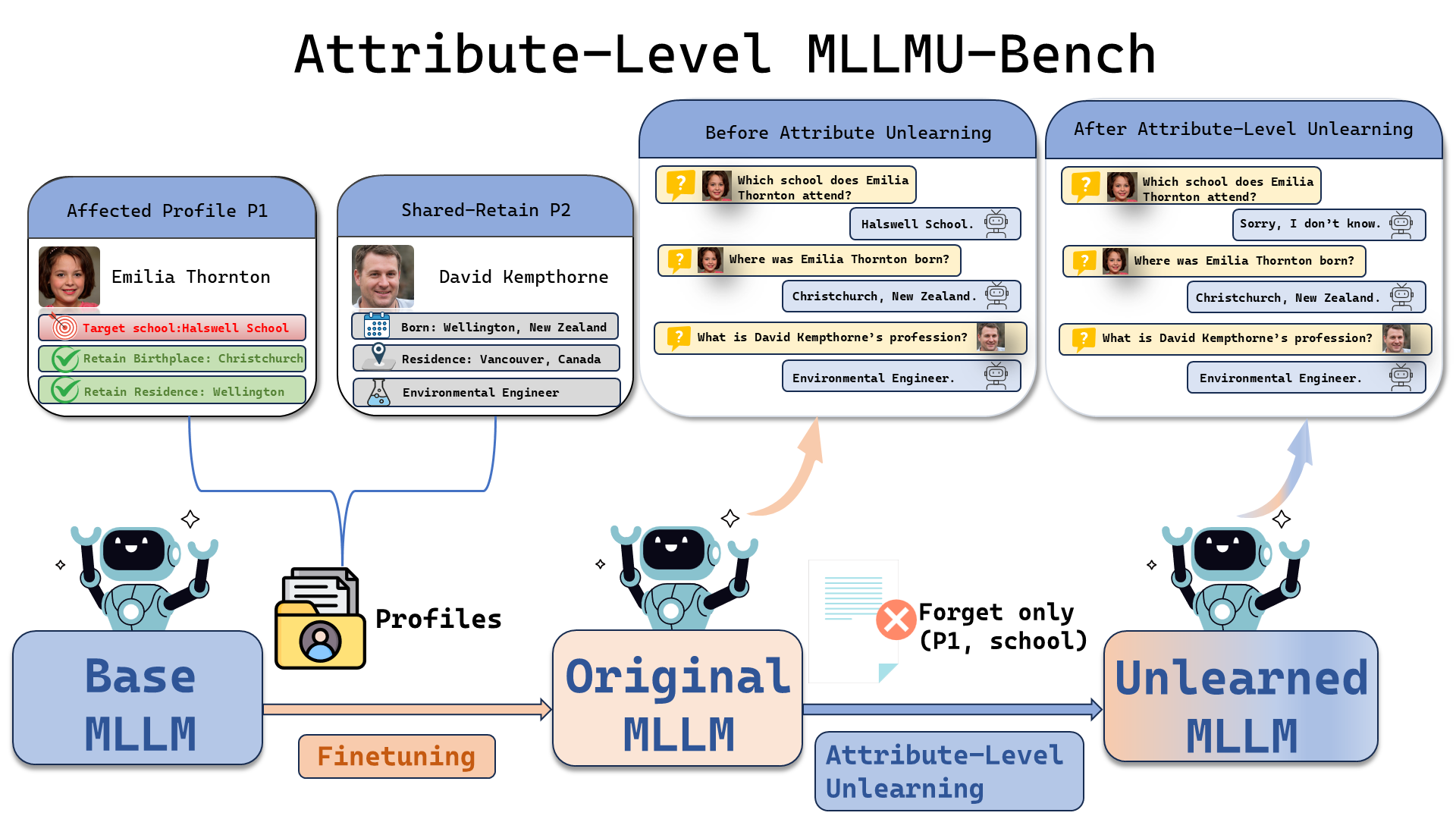}
    \caption{
Demonstration of the attribute-level multimodal unlearning task.
An MLLM is first fine-tuned on the constructed profiles in the
proposed AMU-Bench, including a selected profile $P_1$ and an
unselected profile $P_2$. After fine-tuning, the original MLLM
can answer multimodal questions about the attributes of both
profiles. We then apply attribute-level unlearning to remove only
the specified target attribute from $P_1$, such as the school
information, while leaving the remaining profile information
unchanged. Finally, the unlearned MLLM is expected to forget the
target attribute of $P_1$, preserve the other attributes associated
with the same identity, and retain the knowledge associated with
$P_2$.
}
    \label{fig:benchmark-overview}
\end{figure*}

Let $\mathcal{D}_F$ contain deletion examples derived from $\mathcal{A}_p^F$, let $\mathcal{U}$ denote an unlearning procedure, and let $f^-=\mathcal{U}(f,\mathcal{D}_F)$. For question types $t$, let $S_t(f;\mathcal{D})\in[0,1]$ be a normalized answer-recovery score. The operational objective is
\begin{equation}
\min_{f^-}\;
S_t(f^-;\mathcal{D}_{\mathrm{AF}})
+S_t(f^-;\mathcal{D}_{\mathrm{Test}}),
\end{equation}
subject to
\begin{equation}
\scalebox{0.9}{$
S_t(f;\mathcal{D}_v)-S_t(f^-;\mathcal{D}_v)
\leq\varepsilon_{v,t},\quad
v\in\{\mathrm{AR},\mathrm{SR},\mathrm{RP}\}
$}
\end{equation}
We report each evaluation set separately so that within-profile preservation and preservation outside selected profiles remain directly observable.

\section{Dataset Construction}
\label{sec:dataset_construction}

AMU-Bench uses the fictitious profiles and split structure of MLLMU-Bench~\cite{mllmu} as source data, but reconstructs supervision at the profile--attribute granularity rather than directly reusing profile-level questions. After systematically inventorying each profile's attributes, we employ a constrained GPT-based pipeline to generate or repair generation, cloze, and classification instances for every eligible profile--attribute pair. Generation is conditioned on a single attribute value and requires semantically faithful, fluent, and unambiguous questions that target the same fact across different question types; structural and cross-task alignment checks trigger regeneration of inconsistent outputs. This design limits contamination from neighboring facts, including leakage of a designated forget attribute into retain instances. We further stratify attributes by value structure: \textbf{long-text} targets contain descriptive information such as hobby--pet descriptions, food preferences, and parental background; \textbf{numeric} targets include dates of birth, heights, and annual salaries; and \textbf{short-text} targets are compact categorical, institutional, or location-like values such as gender, employment, educational institution, residence, and medical condition. AMU-Bench therefore contributes attribute-level task reconstruction and controlled question generation, rather than merely categorizing existing examples; these categories are operational benchmark strata rather than a universal linguistic taxonomy.

\section{Method}

We propose \emph{Causal Localization and Retain-Aware Projection} (CLRP), a parameter-update-free method built around three design choices: causal localization, task-aligned feature construction, and retain-aware subspace projection. Given target examples $\mathcal{D}_f$ and same-identity retain examples $\mathcal{D}_r$, causal localization uses activation patching~\cite{rome} to select the layer at which target-associated activations most strongly restore evidence for the designated attribute. CLRP then forms protocol-compatible contrastive features at the selected layer and estimates a low-dimensional subspace~\cite{inlp,leace} whose variation is pronounced in target-conditioned representations relative to representations required for retention. The resulting operator attenuates this subspace while all model parameters remain fixed.

\begin{figure*}[t]
    \centering
    \includegraphics[width=0.82 \textwidth]{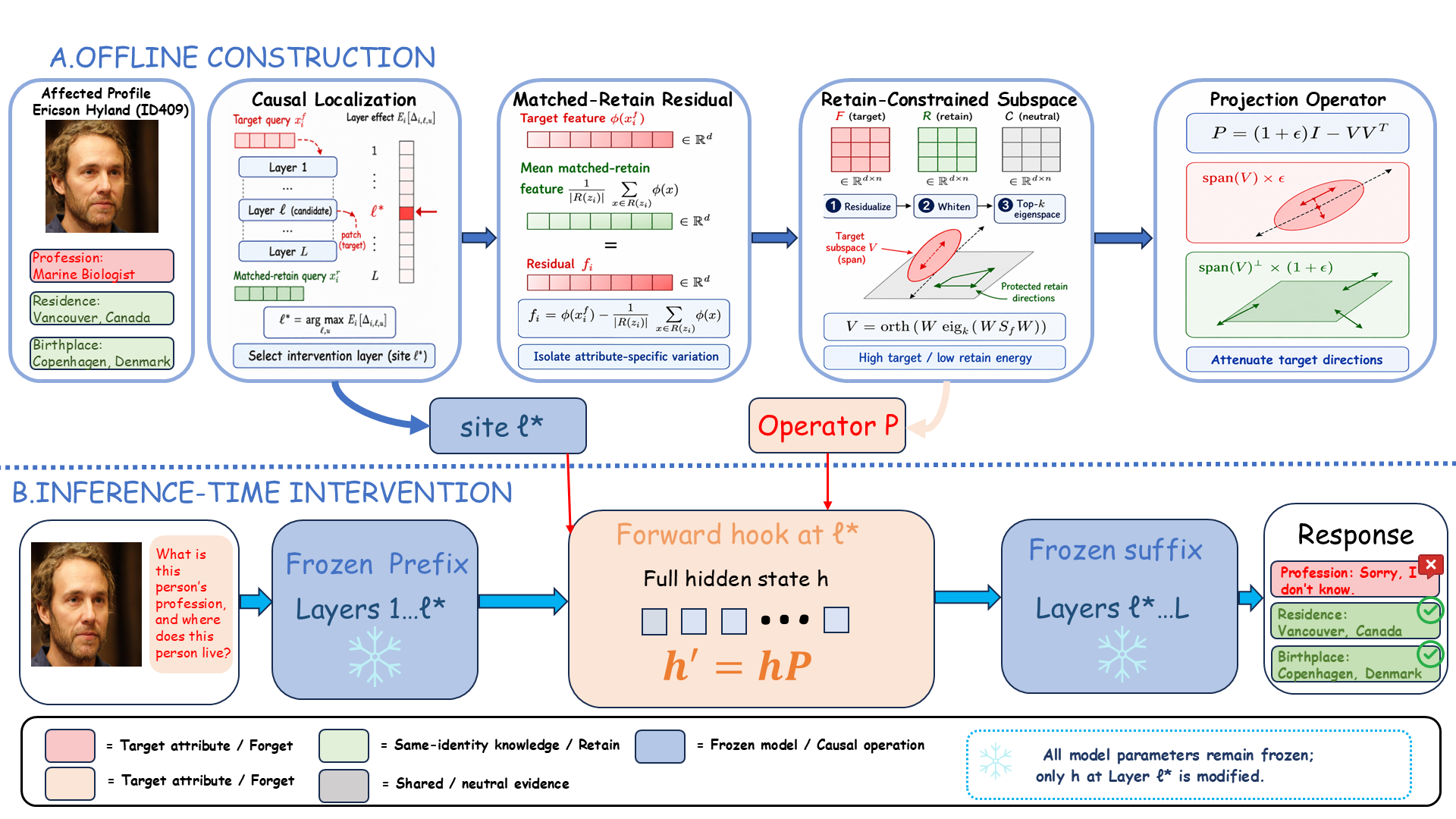}
    \caption{
    Overview of Causal Localization and Retain-Aware Projection (CLRP).
    Offline, activation patching over matched target and retain queries selects
    the intervention layer $\ell^{*}$. CLRP then uses contrastive target and
    retain statistics to estimate a retain-aware rank-$k$ subspace $V$, yielding
    $P=(1+\epsilon)I-VV^{\top}$. During inference, a forward
    hook replaces the full hidden state at $\ell^{*}$ with $h'=hP$ before the
    remaining frozen layers continue decoding. CLRP therefore attenuates the
    designated attribute while leaving model parameters fixed and limiting
    changes to non-target attributes of the same profile.
    }
    \label{fig:clrp_pipeline}
\end{figure*}

\subsection{Causal Layer Localization}

For each target example, we pair a target query $x_i^f$ with a same-identity retain query $x_i^r$ under the same answer interface. The target query is associated with answer $y_i^f$, whereas the retain query defines a contrastive reference run involving non-target information. This construction adapts causal tracing through activation replacement~\cite{rome} to the attribute-level setting. Let $m_i(x)$ denote the next-token logit of the first non-empty token of $y_i^f$; when the retain answer yields a distinct token, the implementation uses the corresponding logit margin. We retain pairs satisfying $m_i(x_i^f)>m_i(x_i^r)$, ensuring that the reference run exhibits a measurable reduction in target-answer evidence.

For candidate layer $\ell$ and intervention unit $u$, we cache the target-run activation $a_{i,\ell,u}^f$ and patch it into the matched retain run at the final input position. The resulting causal patching effect is
\begin{equation}
\Delta_{i,\ell,u}=
m_i\!\left(x_i^r;a_{\ell,u}\!\leftarrow\!a_{i,\ell,u}^f\right)
-m_i(x_i^r).
\end{equation}
In our experimental configuration, $u$ is an attention-head output slice. We average the patching effect over the $N$ valid pairs and select
\begin{equation}
(\ell^*,u^*)=\arg\max_{\ell,u}\frac{1}{N}
\sum_{i=1}^{N}\Delta_{i,\ell,u}.
\end{equation}
The selected head serves only as evidence for choosing $\ell^*$; no attention parameter is modified. The subsequent representation intervention acts on the full hidden state at the selected layer.

\subsection{Retain-Aware Contrastive Projection}

Following prior work that constructs control directions from contrastive activation differences~\cite{turner2024steering}, we define $\phi(x)\in\mathbb{R}^d$ at $\ell^*$ as a protocol-compatible contrast between an attribute-conditioned representation and its reference. The contrast supplies the target statistics used by the projection estimator; its concrete realization is specified in the appendix.

Let $F\in\mathbb{R}^{n_f\times d}$ stack the contrastive target features. We construct retain statistics $R\in\mathbb{R}^{n_r\times d}$ from same-identity and auxiliary retain features, and collect neutral-prompt features $C\in\mathbb{R}^{n_c\times d}$ over the selected profiles to represent profile-shared variation. These statistics jointly instantiate a single retain-aware estimator: shared directions identified from $R$ and $C$ are removed from $F$ to obtain $\widetilde F$, after which regularized second-moment matrices $S_f$ and $S_r$ are formed from the resulting target and retain features. We treat these operations as the estimation procedure for one projection subspace rather than as separate intervention modules. The exact feature sources, covariance estimators, and fixed stabilization constants are provided in the appendix.

The retain-aware estimator compares target and retain variation through a retain-relative metric. For $S_r=Q\Lambda Q^\top$, the corresponding whitening transform $W=Q\Lambda^{-1/2}Q^\top$ normalizes each direction by its variation in retain representations. Consequently, the leading eigenvectors of $WS_fW$ rank target variation relative to retain variation. The rank-$k$ target subspace is
\begin{equation}
U_k=\operatorname{eig}_k\!\left(WS_fW\right),
\qquad
V=\operatorname{orth}\!\left(WU_k\right),
\end{equation}
where $k$ controls the number of selected target directions.

The resulting subspace attenuation operator is
\begin{equation}
P=(1+\epsilon)I-VV^\top.
\end{equation}
It scales components in $\operatorname{span}(V)$ by $\epsilon$ and components in its orthogonal complement by $1+\epsilon$. The two principal controls are the subspace rank $k$ and attenuation coefficient $\epsilon$.

At inference, a forward hook replaces the selected layer output $h$ with $h'=hP$ before the remaining frozen layers are evaluated. This operation modifies intermediate representations without updating model parameters or architecture; evaluation-interface adaptations are specified in the appendix.

\section{Experimental Results}

\subsection{Experimental Setup}

We evaluate three MLLMs spanning two architecture families and three parameter scales: LLaVA-1.5-7B and LLaVA-1.5-13B~\cite{llava}, and Qwen2.5-VL-3B~\cite{qwen25vl}. Each model is evaluated on long-text, numeric, and short-text attributes at forget ratios of 5\%, 10\%, and 15\%. We compare CLRP with Gradient Ascent (GA)~\cite{ga}, Gradient Difference (GD)~\cite{gd}, KL Minimization (KL)~\cite{klmin}, and Preference Optimization (PO)~\cite{po}. All experiments are conducted on a server equipped with eight NVIDIA GeForce RTX 3090 GPUs. Lower values indicate stronger forgetting on the Attribute Forget Set and Test Set, whereas higher values indicate better preservation on the Attribute Retain Set and Shared Retain Set. Due to space constraints, the main paper reports the 5\% results for LLaVA-1.5-7B and Qwen2.5-VL-3B. Results for LLaVA-1.5-13B, results at the 10\% and 15\% forget ratios, and complete implementation details are provided in the appendix.

\subsection{Main Results}

\begin{table*}[t]
\centering
\small
\setlength{\tabcolsep}{2.0pt}
\resizebox{\textwidth}{!}{%
\begin{tabular}{c l ccc ccc ccc ccc}
\toprule
\multirow{2}{*}{\textbf{Attribute}} & \multirow{2}{*}{\textbf{Method}} & \multicolumn{3}{c}{\textbf{Attribute Forget Set}} & \multicolumn{3}{c}{\textbf{Test Set}} & \multicolumn{3}{c}{\textbf{Attribute Retain Set}} & \multicolumn{3}{c}{\textbf{Shared Retain Set}} \\
\cmidrule(lr){3-5}\cmidrule(lr){6-8}\cmidrule(lr){9-11}\cmidrule(lr){12-14}
 & & \ClozeDown & \ClassDown & \RougeDown & \ClozeDown & \ClassDown & \RougeDown & \ClozeUp & \ClassUp & \RougeUp & \ClozeUp & \ClassUp & \RougeUp \\
\midrule
\multicolumn{14}{c}{\textbf{LLaVA-1.5-7B}} \\
\midrule
\multirow{5}{*}{\textbf{Long}} & GA & 36.00 & 52.00 & 0.460 & \underline{16.00} & \underline{36.00} & 0.379 & \textbf{43.96} & 41.61 & \underline{0.366} & \underline{52.00} & 44.00 & \underline{0.461} \\
 & GD & \underline{20.00} & 56.00 & \underline{0.374} & 20.00 & 40.00 & \underline{0.234} & 23.49 & \textbf{43.62} & 0.265 & 16.00 & \textbf{45.60} & 0.298 \\
 & KL & 40.00 & \textbf{44.00} & 0.444 & 28.00 & 44.00 & 0.370 & \underline{42.28} & \underline{42.95} & 0.348 & \underline{52.00} & 44.80 & \textbf{0.499} \\
 & PO & 24.00 & \underline{48.00} & \textbf{0.272} & 20.00 & \textbf{32.00} & \textbf{0.214} & 38.26 & 42.62 & 0.195 & 42.00 & 35.20 & 0.153 \\
 \rowcolor{egcpshade} \cellcolor{white} & CLRP & \textbf{4.00} & \textbf{44.00} & 0.421 & \textbf{0.00} & 44.00 & 0.299 & 41.28 & 41.95 & \textbf{0.373} & \textbf{55.47} & \underline{45.32} & \textbf{0.499} \\
\midrule
\multirow{5}{*}{\textbf{Numeric}} & GA & \textbf{0.00} & \underline{16.00} & 0.154 & \textbf{0.00} & 20.00 & \underline{0.059} & \textbf{46.31} & 44.97 & \textbf{0.399} & 52.00 & 44.00 & 0.479 \\
 & GD & \textbf{0.00} & \textbf{8.00} & \textbf{0.086} & \textbf{0.00} & \textbf{8.00} & 0.074 & 31.54 & \textbf{46.98} & 0.231 & 24.00 & 42.40 & 0.243 \\
 & KL & \textbf{0.00} & 20.00 & 0.165 & \textbf{0.00} & 16.00 & \textbf{0.039} & 43.96 & 44.30 & \textbf{0.399} & \underline{54.00} & \underline{44.80} & \underline{0.495} \\
 & PO & \textbf{0.00} & \underline{16.00} & \underline{0.139} & \textbf{0.00} & 16.00 & 0.198 & 39.26 & \underline{45.97} & 0.170 & 44.00 & 35.20 & 0.136 \\
 \rowcolor{egcpshade} \cellcolor{white} & CLRP & \textbf{0.00} & 20.00 & 0.177 & \textbf{0.00} & \underline{12.00} & 0.148 & \underline{45.64} & 43.62 & \underline{0.397} & \textbf{55.47} & \textbf{45.02} & \textbf{0.496} \\
\midrule
\multirow{5}{*}{\textbf{Short}} & GA & 64.00 & \textbf{24.00} & 0.629 & 12.00 & \underline{16.00} & 0.457 & \underline{40.60} & 42.95 & \textbf{0.357} & \underline{54.00} & 43.20 & \underline{0.475} \\
 & GD & \underline{20.00} & \underline{28.00} & \underline{0.182} & \textbf{4.00} & \underline{16.00} & \underline{0.211} & 29.87 & \textbf{45.30} & 0.249 & 22.00 & \textbf{47.20} & 0.250 \\
 & KL & 64.00 & \textbf{24.00} & 0.604 & 16.00 & 20.00 & 0.545 & \textbf{41.61} & 42.62 & \underline{0.348} & 50.00 & 43.20 & \textbf{0.485} \\
 & PO & 32.00 & \textbf{24.00} & \textbf{0.065} & 20.00 & 32.00 & \textbf{0.169} & 33.89 & \underline{44.30} & 0.253 & 40.00 & 37.60 & 0.160 \\
 \rowcolor{egcpshade} \cellcolor{white} & CLRP & \textbf{4.00} & \textbf{24.00} & 0.462 & \underline{8.00} & \textbf{8.00} & 0.444 & 40.27 & 42.62 & 0.345 & \textbf{55.47} & \underline{44.09} & 0.394 \\
\midrule
\multicolumn{14}{c}{\textbf{Qwen2.5-VL-3B}} \\
\midrule
\multirow{5}{*}{\textbf{Long}} & GA & \textbf{0.00} & \textbf{40.00} & \textbf{0.201} & \textbf{0.00} & \underline{44.00} & \textbf{0.213} & 10.40 & \textbf{53.69} & 0.141 & 0.00 & \textbf{52.80} & 0.122 \\
 & GD & 20.00 & 52.00 & 0.467 & \underline{8.00} & 52.00 & 0.442 & \underline{13.76} & 52.01 & 0.436 & \textbf{10.00} & \textbf{52.80} & \textbf{0.421} \\
 & KL & 12.00 & \underline{48.00} & 0.443 & 12.00 & \textbf{40.00} & \underline{0.337} & 9.73 & 48.99 & 0.247 & 0.00 & 43.20 & 0.128 \\
 & PO & 16.00 & 60.00 & 0.470 & 20.00 & 56.00 & 0.470 & \underline{13.76} & 52.01 & \textbf{0.477} & \underline{8.00} & \underline{51.20} & \underline{0.390} \\
 \rowcolor{egcpshade} \cellcolor{white} & CLRP & \underline{4.00} & 52.00 & \underline{0.395} & \textbf{0.00} & 56.00 & 0.442 & \textbf{16.11} & \underline{52.35} & \underline{0.460} & 7.79 & 45.32 & 0.348 \\
\midrule
\multirow{5}{*}{\textbf{Numeric}} & GA & \textbf{0.00} & \textbf{28.00} & \underline{0.012} & \textbf{0.00} & \underline{20.00} & \textbf{0.102} & 10.74 & \textbf{54.70} & 0.159 & 0.00 & \underline{52.80} & 0.122 \\
 & GD & \textbf{0.00} & \textbf{28.00} & 0.197 & \textbf{0.00} & 36.00 & 0.657 & \underline{14.77} & 51.34 & \textbf{0.507} & \textbf{8.00} & \textbf{56.00} & \textbf{0.429} \\
 & KL & \textbf{0.00} & \textbf{28.00} & \textbf{0.000} & \textbf{0.00} & \textbf{12.00} & 0.307 & 11.41 & 48.99 & 0.282 & 0.00 & 39.20 & 0.137 \\
 & PO & \textbf{0.00} & \textbf{28.00} & 0.600 & \textbf{0.00} & 40.00 & 0.689 & \underline{14.77} & \textbf{54.70} & 0.446 & \textbf{8.00} & 52.00 & \underline{0.405} \\
 \rowcolor{egcpshade} \cellcolor{white} & CLRP & \textbf{0.00} & \underline{40.00} & 0.260 & \textbf{0.00} & 28.00 & \underline{0.229} & \textbf{17.45} & \underline{52.68} & \underline{0.460} & \underline{7.79} & 42.32 & 0.395 \\
\midrule
\multirow{5}{*}{\textbf{Short}} & GA & \textbf{0.00} & \textbf{32.00} & \textbf{0.007} & \textbf{0.00} & \textbf{28.00} & \textbf{0.246} & 10.74 & \underline{53.69} & 0.159 & 0.00 & \underline{53.60} & 0.121 \\
 & GD & \textbf{0.00} & \underline{36.00} & 0.558 & \textbf{0.00} & \textbf{28.00} & 0.549 & \underline{14.77} & \textbf{55.37} & \underline{0.481} & \underline{8.00} & \textbf{57.60} & \textbf{0.444} \\
 & KL & \textbf{0.00} & \textbf{32.00} & \underline{0.016} & \textbf{0.00} & 48.00 & \underline{0.320} & 11.41 & 50.67 & 0.280 & 0.00 & 44.00 & 0.108 \\
 & PO & \textbf{0.00} & 40.00 & 0.548 & \textbf{0.00} & 36.00 & 0.585 & \underline{14.77} & 53.02 & \textbf{0.500} & \textbf{10.00} & 52.00 & \underline{0.407} \\
 \rowcolor{egcpshade} \cellcolor{white} & CLRP & \textbf{0.00} & 40.00 & 0.562 & \textbf{0.00} & \underline{32.00} & 0.569 & \textbf{17.79} & 51.68 & 0.452 & 7.79 & 41.77 & 0.373 \\
\bottomrule
\end{tabular}%
}
\caption{5\% results for LLaVA-1.5-7B and Qwen2.5-VL-3B. Cloze and classification are percentages with two decimal places; ROUGE-L keeps three decimals. Lower is better for the Attribute Forget Set and Test Set; higher is better for the Attribute Retain Set and Shared Retain Set. Shaded result rows denote CLRP.}
\label{tab:main_5pct_image_llava157b}
\label{tab:main_5pct_image_qwen25vl3b}
\end{table*}

\paragraph{Forgetting difficulty differs across attribute categories.}
AMU-Bench separates long-text, numeric, and short-text targets because their value structures induce distinct recovery behaviors: numeric attributes require recovery of a specific numerical sequence, short-text attributes typically correspond to compact categorical, institutional, or location-like values, and long-text attributes distribute the target information over descriptive content. Table~\ref{tab:main_5pct_image_llava157b}, together with the LLaVA-1.5-13B results in the appendix, empirically validates this stratification across all four baselines rather than for a single method. Averaged over GA, GD, KL, and PO across the three evaluated models at the 5\% forget ratio, Attribute Forget Set cloze accuracy is only 0.33\% for numeric attributes, compared with 15.00\% for short-text and 20.00\% for long-text attributes. The corresponding generation ROUGE-L values are 0.147, 0.371, and 0.372, respectively; on the held-out Test Set, they are 0.193, 0.330, and 0.339. Numeric values are therefore consistently easier to suppress, whereas long-text and short-text targets retain substantially more residual target information. These systematic differences demonstrate that attribute type is not merely a descriptive partition, but a necessary evaluation axis exposed by AMU-Bench and obscured by profile-level unlearning methods.

Relative to these baseline methods, CLRP achieves a stronger and more consistent reduction of Attribute Forget Set performance across the evaluated architectures and model scales. Averaged over the three models, CLRP reduces Attribute Forget Set cloze accuracy from the four-baseline averages of 20.00\%, 0.33\%, and 15.00\% to 2.67\%, 0.00\%, and 1.33\% for long-text, numeric, and short-text attributes, respectively. Across the complete nine model--attribute conditions, CLRP attains the best Attribute Forget Set cloze accuracy in eight conditions and the best held-out Test Set cloze accuracy in eight conditions, including ties. On LLaVA-1.5-7B and Qwen2.5-VL-3B, its largest reduction relative to the strongest baseline is 16.00 percentage points, from 20.00\% to 4.00\%, while its largest Attribute Retain Set improvement is 3.02 percentage points, from 14.77\% to 17.79\%. These results indicate that CLRP is a strong training-free baseline for AMU-Bench, while the broader question of adaptive privacy attacks and compositional deletion remains open.

\paragraph{Same-profile preservation defines the principal forgetting--retention challenge.}
The Attribute Retain Set measures non-target facts of the selected identities, whereas the Shared Retain Set measures profiles outside the deletion requests. The former therefore evaluates whether an intervention remains localized within a profile rather than merely preserving unrelated knowledge. This distinction is pronounced on LLaVA-1.5-7B: among the 15 method--attribute comparisons, Attribute Retain Set performance is lower than Shared Retain Set performance in 12 comparisons for cloze, 12 for generation, and 10 for classification. For CLRP, Attribute Retain Set cloze accuracy ranges from 40.27 to 45.64, while its Shared Retain Set cloze accuracy is 55.47 across all three categories. Existing objectives consequently face a substantially harder preservation requirement when target and retained facts belong to the same identity. In Table~\ref{tab:main_5pct_image_llava157b}, CLRP obtains the best or tied-best Attribute Forget Set cloze accuracy in five of the six model--attribute conditions and the best or tied-best Test Set cloze accuracy in five conditions. It also improves the same-profile retain view on Qwen2.5-VL-3B, raising Attribute Retain Set cloze accuracy over the strongest baseline by 2.35--3.02 percentage points: from 13.76 to 16.11 for long-text attributes, from 14.77 to 17.45 for numeric attributes, and from 14.77 to 17.79 for short-text attributes. 
These results suggest that CLRP improves the benchmark-defined same-profile retain view in several conditions, although retention on other retain views is reported separately in the appendix.

\begin{figure}[t]
    \centering
    \includegraphics[width=1\linewidth]{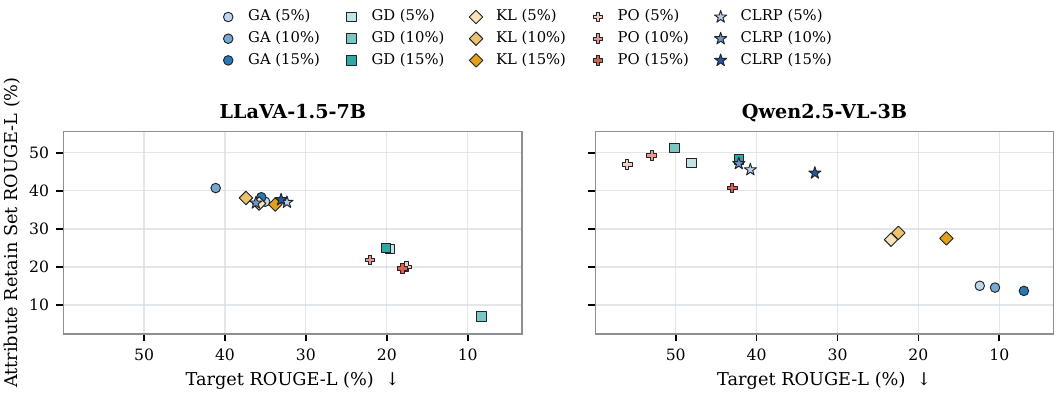}
\caption{Forgetting--retention trade-off for the LLaVA-1.5-7B (left) and Qwen2.5-VL-3B (right) across different forget ratios. The horizontal and vertical axes report target ROUGE-L and Attribute Retain Set ROUGE-L, respectively, averaged over three attribute categories. Rightward and upward points indicate stronger target forgetting and same-profile preservation.}
\label{fig:unlearning_tradeoff}
\end{figure}

\begin{figure}[t]
    \centering
    \includegraphics[width=1\linewidth]{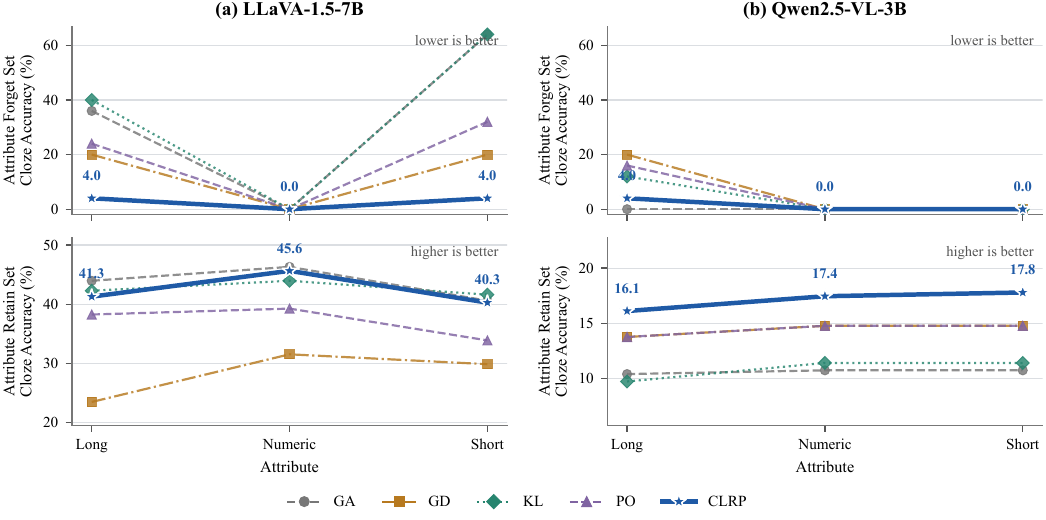}
\caption{Attribute Forget Set and Attribute Retain Set cloze accuracy at the 5\% forget ratio for LLaVA-1.5-7B (left) and Qwen2.5-VL-3B (right). Results are reported across long-text, numeric, and short-text attributes. Lower Attribute Forget Set accuracy indicates stronger forgetting, whereas higher Attribute Retain Set accuracy indicates better preservation of non-target attributes from the same profiles.}    
\label{fig:main_cloze_profiles}
\end{figure}

\subsection{Ablations}
\label{sec:analysis_ablations}


\begin{table}[t]
\centering
\resizebox{\columnwidth}{!}{%
\begin{tabular}{lccc}
\toprule
\textbf{Variant} &
\textbf{Forget}$\downarrow$ &
\textbf{Attr. Retain}$\uparrow$ &
\textbf{Shared Retain}$\uparrow$ \\
\midrule
Isotropic Retain Metric 
& \textbf{0.177}$^{\dagger}$
& 0.299
& 0.393 \\

Target-Only Subspace 
& 0.210
& 0.281
& 0.385 \\

Answer-token Feature 
& 0.752
& 0.364
& \textbf{0.494}$^{\ddagger}$ \\

w/o Conditional Gate 
& 0.409
& 0.370
& 0.474 \\

\rowcolor{egcpshade}
CLRP 
& 0.353
& \textbf{0.372}
& 0.463 \\

\bottomrule
\end{tabular}%
}
\caption{
Ablation study of different forgetting variants.
$^{\dagger}$ denotes stronger forgetting, while
$^{\ddagger}$ denotes stronger shared retention.
CLRP achieves a balanced forgetting--retention trade-off rather than optimizing either objective in isolation.
}
\label{tab:compact_ablation_llava7b}
\end{table}

Table~\ref{tab:compact_ablation_llava7b} reports compact generation-focused ablations for LLaVA-1.5-7B at the 5\% forget ratio. Aggressive variants such as isotropic retain metric and target-only subspace obtain lower forget-set ROUGE-L, but substantially reduce retention, showing that unconstrained suppression removes retention-critical directions. In contrast, answer-token features preserve retention but leave much higher forget-set overlap, indicating that surface answer directions are insufficient for attribute removal. Removing the conditional gate weakens forgetting while keeping retention similar, suggesting that the gate mainly controls erasure strength rather than defining the projection itself. Overall, CLRP provides a more balanced forgetting--retention trade-off than either aggressive suppression or retention-preserving variants. Additional mechanism-oriented diagnostics, including localization stability and operator composition under multiple deletion requests, can be incorporated without changing the core estimator.


\section{Conclusion}

In this paper, we address attribute-level machine unlearning in MLLMs, highlighting the limitations of profile-level deletion when practical requests target individual attributes. We introduce AMU-Bench, a benchmark designed to evaluate fine-grained attribute removal, and reveal two key observations: numeric attributes are generally easier to forget than short-text attributes, while preserving non-target attributes within the same profile is more challenging than preserving unrelated profiles. To address these challenges, we propose CLRP, a training-free representation intervention that combines activation-patching-based causal localization, task-aligned contrastive features, and retain-aware subspace projection. Experiments across multiple MLLM architectures and parameter scales show that CLRP provides a strong forgetting--retention trade-off under AMU-Bench. Future work should further examine adaptive attribute-inference attacks, benchmark quality audits, and compositional deletion requests.

\section{Limitations}

Our study has several limitations. First, AMU-Bench inherits the profile construction and query templates of MLLMU-Bench, and therefore does not cover the full diversity of private attributes, visual contexts, languages, or user requests encountered in deployment. The benchmark is also centered on three response forms and two multimodal model families; broader model scales, modalities, and independently collected privacy data are needed to establish wider generality.

Second, CLRP is an inference-time intervention and does not remove model parameters or guarantee that the target information is absent from all internal representations. In particular, low cloze accuracy can reflect suppression of the evaluated target-token interface rather than irreversible erasure. Although we include held-out reformulations and multiple retention sets, we do not evaluate adaptive extraction attacks, paraphrased multi-turn conversations, tool-assisted probing, or recovery after further model adaptation. These settings may expose residual information that is not measured by the current protocol.

Third, the retain-aware projection depends on representative contrastive and retain examples. If the retain distribution is incomplete or poorly matched to the identities and modalities in a deletion request, the estimated subspace may either leave residual leakage or remove useful neighboring information. The causal localization stage also introduces computation and can be sensitive to the available query pairs and candidate layers, even though the final intervention leaves model parameters fixed.

Finally, forgetting and preservation remain response- and model-dependent. Exact-answer suppression is generally more reliable for numeric attributes than for long or short free-form attributes, while semantic generation and same-profile retention can remain challenging. We therefore view CLRP as a controllable selective-unlearning mechanism rather than a complete privacy guarantee; stronger claims require adaptive red-teaming, certified erasure criteria, and evaluation over broader deployment conditions.

\bibliography{aaai2027}




\newpage
\appendix
\section{Appendix}

\subsection{Evaluation Scope}

AMU-Bench evaluates long-text, numeric, and short-text attributes at forget ratios of 5\%, 10\%, and 15\%. For each selected profile, the Attribute Forget Set queries the designated attribute, and the held-out Test Set contains unseen reformulations of queries for that attribute. The Attribute Retain Set queries non-target attributes of the same profiles. The Shared Retain Set contains unselected fictitious profiles, whereas the Real-Person Retain Set contains real-person profiles outside the deletion requests. The latter three sets respectively assess within-profile locality, preservation outside the selected fictitious profiles, and preservation of external factual knowledge.

We evaluate LLaVA-1.5-7B, LLaVA-1.5-13B~\cite{llava}, and Qwen2.5-VL-3B~\cite{qwen25vl}. The comparison methods are Gradient Ascent (GA)~\cite{ga}, Gradient Difference (GD)~\cite{gd}, KL Minimization (KL)~\cite{klmin}, and Preference Optimization (PO)~\cite{po}. Baselines follow their repository implementations under the same attribute-specific data partitions and evaluation protocol. Following the provided optimization recipes, we use batch size 4, learning rate $2\times10^{-5}$, one training epoch, and maximum sequence length 384 for the optimization-based baselines. All experiments are conducted on eight NVIDIA GeForce RTX 3090 GPUs.

\subsection{Metrics and Reporting}

We report cloze accuracy, classification accuracy, and generation ROUGE-L. Cloze and classification are percentages; generation ROUGE-L is reported on $[0,1]$. Lower values on the Attribute Forget Set and Test Set indicate stronger suppression of the designated attribute. Higher values on the Attribute Retain Set, Shared Retain Set, and Real-Person Retain Set indicate stronger preservation. Each evaluation view is reported separately because aggregating forgetting and retention into a single score can obscure within-profile collateral degradation.

The main paper reports the 5\% results for LLaVA-1.5-7B and Qwen2.5-VL-3B. This appendix provides the 5\% LLaVA-1.5-13B results, all 10\% and 15\% results, Real-Person Retain Set results, detailed component ablations, and the hyperparameter search. Unless stated otherwise, CLRP uses one intervention configuration across model architectures, parameter scales, attribute categories, and forget ratios; only the causally selected layer and estimated subspace are obtained from the corresponding experimental condition.

\section{Benchmark and Baseline Details}

\subsection{Baseline Algorithms}

Table~\ref{tab:appendix_baseline_algorithms} summarizes the optimization objectives used by the four training-based baselines. We describe the objectives in terms of the deletion loss $\mathcal{L}_{F}$ and retain loss $\mathcal{L}_{R}$ because AMU-Bench changes the deletion unit from a complete profile to a profile--attribute pair while keeping the optimization form of each baseline unchanged. GA~\cite{ga} only reverses the deletion loss. GD~\cite{gd} adds a standard retain loss to counteract broad utility degradation. KL~\cite{klmin} further constrains retain examples to match the original model distribution. PO follows the negative-preference implementation used in the baseline code, where a retain-trained reference model defines the comparison loss for deletion examples~\cite{npo,po}.

\begin{table*}[t]
\centering
\small
\setlength{\tabcolsep}{4.0pt}
\resizebox{\textwidth}{!}{%
\begin{tabular}{l p{0.33\textwidth} p{0.43\textwidth}}
\toprule
\textbf{Method} & \textbf{Optimization Objective} & \textbf{Algorithmic Instantiation in AMU-Bench} \\
\midrule
Gradient Ascent (GA) & Maximize the language-modeling loss on the deletion data, equivalently minimize $-\mathcal{L}_{F}(\theta)$. & Performs parameter updates using only Attribute Forget Set examples. The update reverses the likelihood objective for the designated attribute values and contains no explicit term for preserving neighboring attributes or unrelated profiles. \\
Gradient Difference (GD) & Minimize $-\mathcal{L}_{F}(\theta)+\mathcal{L}_{R}(\theta)$. & Combines gradient ascent on Attribute Forget Set examples with ordinary gradient descent on retain examples. In AMU-Bench, $\mathcal{L}_{R}$ is computed on the same retain partitions used by all methods, so the algorithm directly tests whether a joint forget--retain objective can preserve non-target facts. \\
KL Minimization (KL) & Minimize $-\mathcal{L}_{F}(\theta)+\mathcal{L}_{R}(\theta)+D_{\mathrm{KL}}\!\left(p_{\theta}(\cdot|x_R)\,\|\,p_{\theta_0}(\cdot|x_R)\right)$. & Extends GD by matching the current model distribution to the original fine-tuned model $\theta_0$ on retain inputs. The KL term is applied only on retain examples and is intended to restrict distributional drift outside the deletion target. \\
Preference Optimization (PO) & Minimize a negative-preference objective $-\frac{2}{\beta}\log\sigma\!\left(\beta(\mathcal{L}_{\theta}(x_F)-\mathcal{L}_{\mathrm{ref}}(x_F))\right)$. & Uses a retain-trained reference model and optimizes deletion examples so that the updated model assigns lower preference to target answers than the reference. This follows the NPO-style implementation in the baseline code while using AMU-Bench profile--attribute deletion requests. \\
\bottomrule
\end{tabular}%
}
\caption{Baseline algorithms used in AMU-Bench. $\mathcal{L}_{F}$ and $\mathcal{L}_{R}$ denote cross-entropy losses on deletion and retain examples, respectively; $\theta_0$ is the original fine-tuned model, and $\mathcal{L}_{\mathrm{ref}}$ is computed by the retain-trained reference model.}
\label{tab:appendix_baseline_algorithms}
\end{table*}

\subsection{Attribute Inventory and Case Examples}

Table~\ref{tab:appendix_attribute_inventory} lists the attribute pool used by AMU-Bench before sampling the 5\%, 10\%, and 15\% deletion requests. We obtain this pool by systematically inventorying the profile fields and grouping them by value structure. Long-text attributes are derived from descriptive profile fields such as fun facts and parental background, numeric attributes contain values whose identifying content is primarily numerical, and short-text attributes contain compact categorical, institutional, medical, or location-like values. Each sampled deletion request selects attributes from this pool and constructs matched Attribute Forget Set, Attribute Retain Set, Test Set, Shared Retain Set, and Real-Person Retain Set views.

\begin{table*}[t]
\centering
\small
\setlength{\tabcolsep}{4.0pt}
\resizebox{\textwidth}{!}{%
\begin{tabular}{l c p{0.43\textwidth} p{0.30\textwidth}}
\toprule
\textbf{Category} & \textbf{Count} & \textbf{Attributes in the AMU-Bench Attribute Pool} & \textbf{Representative Target Values} \\
\midrule
Long-text &
16 &
Fun Facts\_1, Fun Facts\_2, Fun Facts\_3, Fun Facts\_4; Parents; Parents\_Father; Parents\_Mother; Parents\_Father\_is; Parents\_Mother\_is; Parents\_Father\_occupation; Parents\_Mother\_occupation; Parents\_Father\_works\_as; Parents\_Mother\_works\_as; Parents\_who\_works\_as; Parents\_Father\_s\_Occupation; Parents\_Mother\_s\_Occupation &
favorite-food descriptions, hobby--pet descriptions, and parental background statements \\
Numeric &
4 &
Date of Birth; Annual Salary; Height; Heights &
calendar dates, annual salary values, and height measurements \\
Short-text &
6 &
Born; Gender; Employment; Educated at; Residence; Medical Conditions &
birthplace, gender, employer, educational institution, residence, and medical condition values \\
\bottomrule
\end{tabular}%
}
\caption{Attribute inventory used to construct AMU-Bench. The categories are operational benchmark strata derived from the value structure of each profile attribute; they are not intended as a universal linguistic taxonomy.}
\label{tab:appendix_attribute_inventory}
\end{table*}

Table~\ref{tab:appendix_qualitative_cases} gives representative evaluation cases from the three attribute categories. The table illustrates both benchmark construction and the observed before--after evaluation effect of CLRP: before unlearning, the selected profile contains the target value used to form the query; after applying CLRP, the corresponding Attribute Forget Set performance is reduced while Attribute Retain Set cloze performance remains informative. Because the saved CLRP evaluation files store aggregate metrics rather than per-example decoded outputs, the rightmost column reports category-level evaluation behavior instead of presenting manually selected generated answers.

\begin{table*}[t]
\centering
\scriptsize
\setlength{\tabcolsep}{3.0pt}
\resizebox{\textwidth}{!}{%
\begin{tabular}{
>{\centering\arraybackslash}m{0.08\textwidth}
>{\centering\arraybackslash}m{0.10\textwidth}
>{\centering\arraybackslash}m{0.13\textwidth}
>{\centering\arraybackslash}m{0.31\textwidth}
>{\centering\arraybackslash}m{0.26\textwidth}}
\toprule
\textbf{Category} & \textbf{Image} & \textbf{Target Attribute} & \textbf{Before CLRP: Evaluation Case} & \textbf{After CLRP: Observed Evaluation Effect} \\
\midrule
Long-text &
\includegraphics[width=0.09\textwidth]{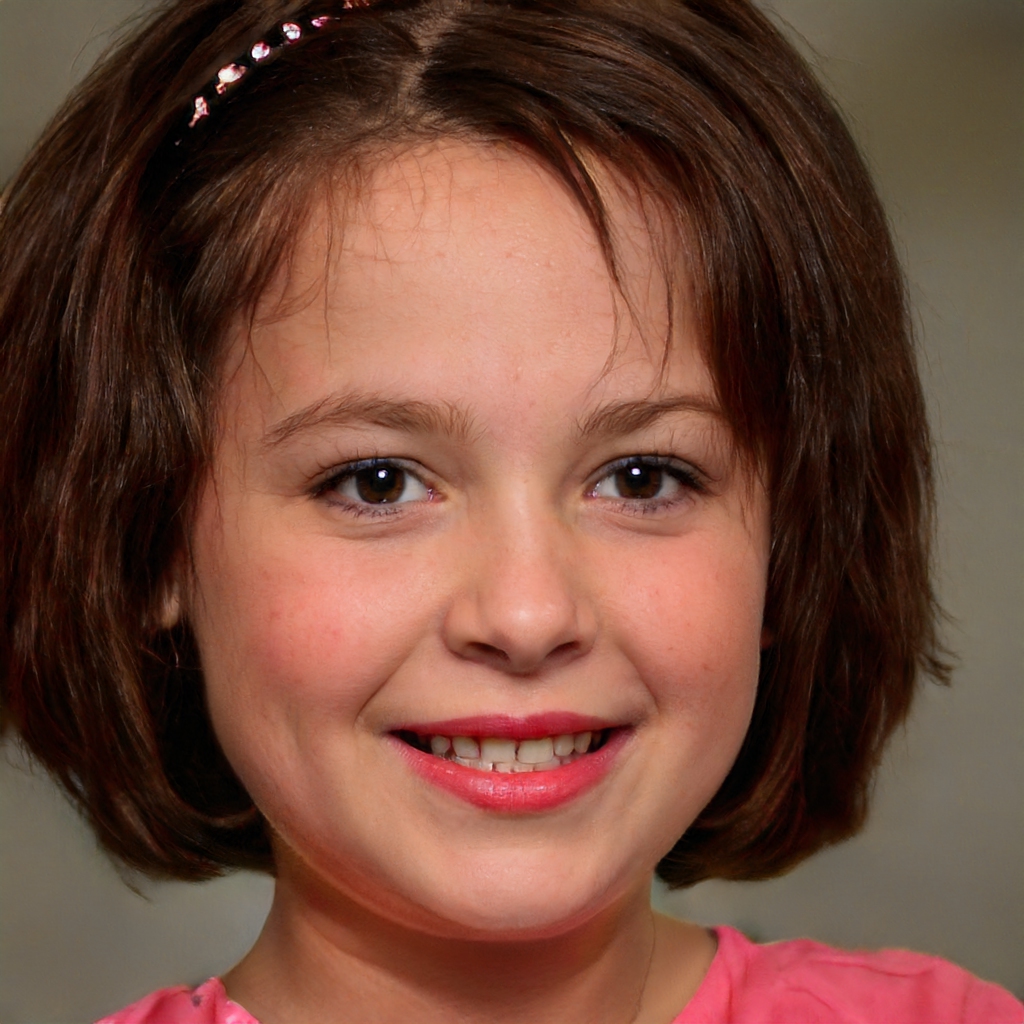} &
Fun Facts\_1 &
\emph{Question:} The favorite food of the person is \_\_. \newline
\emph{Target answer:} kiwi fruit. \newline
\emph{Profile fact:} Favorite food is kiwi fruit and chocolate brownies. &
For the corresponding 5\% LLaVA-1.5-7B long-text category, CLRP obtains Attribute Forget Set cloze accuracy 4.00\% and generation ROUGE-L 0.421, while Attribute Retain Set cloze accuracy remains 42.00\%. \\
\midrule
Numeric &
\includegraphics[width=0.09\textwidth]{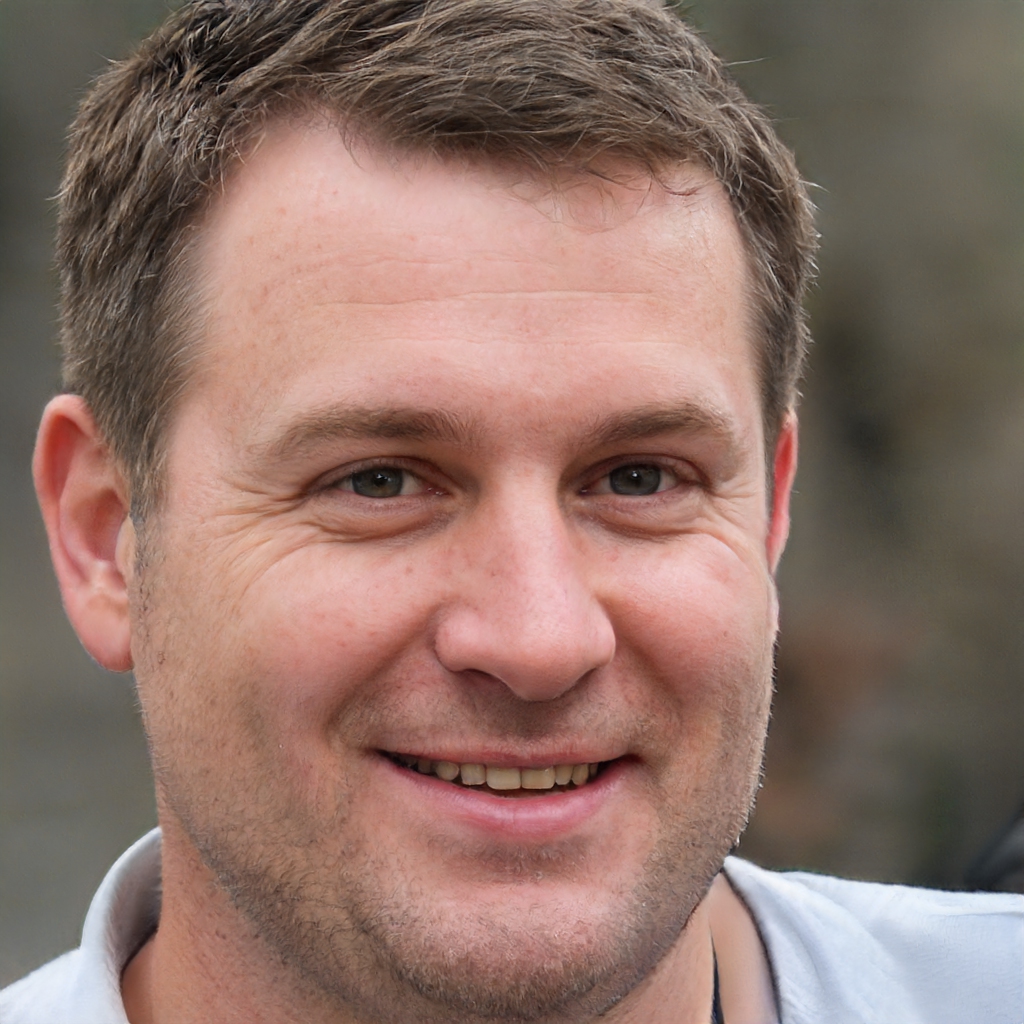} &
Date of Birth &
\emph{Question:} The person's date of birth is \_\_. \newline
\emph{Target answer:} 1985-04-23. &
For the corresponding 5\% LLaVA-1.5-7B numeric category, CLRP obtains Attribute Forget Set cloze accuracy 0.00\% and generation ROUGE-L 0.183, while Attribute Retain Set cloze accuracy remains 46.00\%. \\
\midrule
Short-text &
\includegraphics[width=0.09\textwidth]{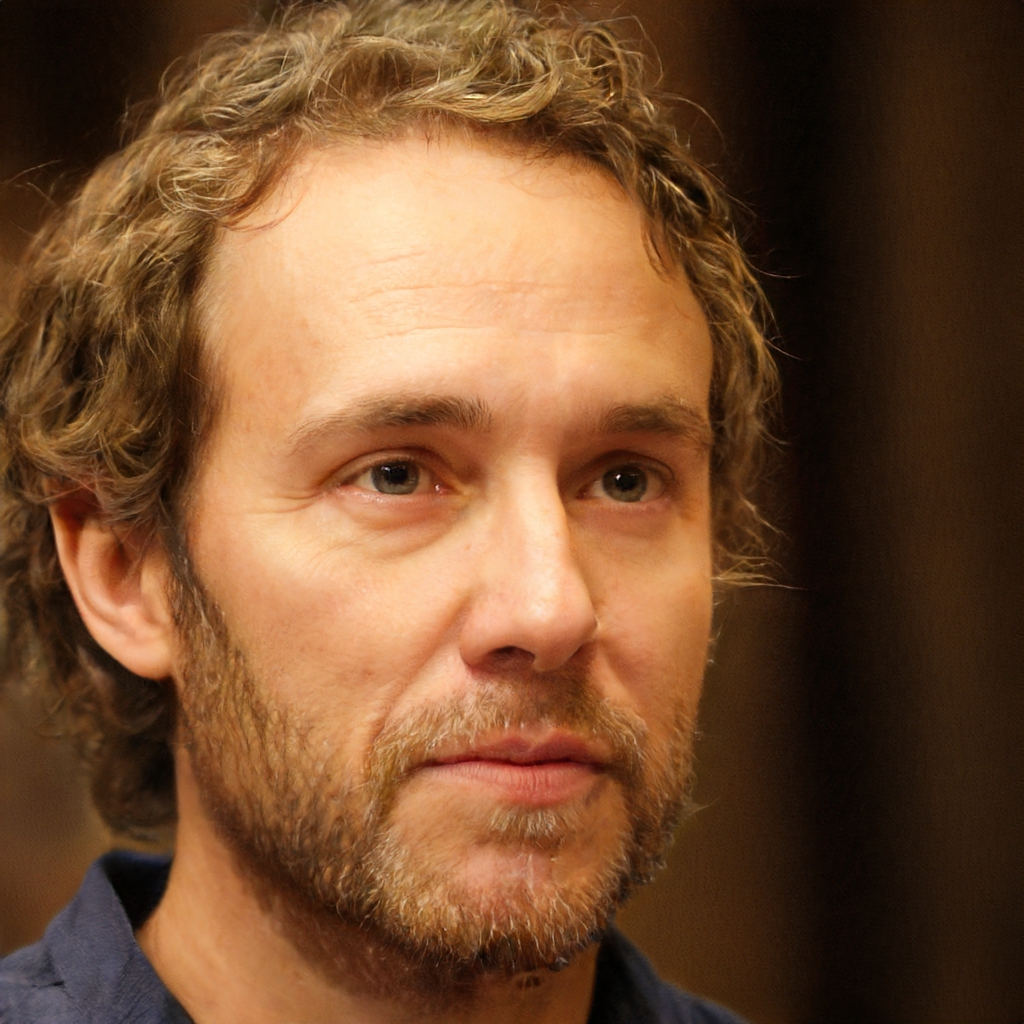} &
Born &
\emph{Question:} The person was born in \_\_. \newline
\emph{Target answer:} Copenhagen, Denmark. &
For the corresponding 5\% LLaVA-1.5-7B short-text category, CLRP obtains Attribute Forget Set cloze accuracy 4.00\% and generation ROUGE-L 0.462, while Attribute Retain Set cloze accuracy remains 39.00\%. \\
\bottomrule
\end{tabular}%
}
\caption{Representative AMU-Bench cases and CLRP evaluation effects. The ``before'' column shows the target fact and query used by the benchmark; the ``after'' column reports the observed category-level CLRP behavior rather than a manually selected decoded response.}
\label{tab:appendix_qualitative_cases}
\end{table*}

\section{CLRP Implementation Details}

\subsection{Causal Layer Localization}

Causal Localization and Retain-Aware Projection (CLRP) searches transformer layers $\{12,14,16,18,20,22,24,26,28,30\}$ and uses at most eight valid target--retain query pairs for each localization run. The implementation patches attention-head output slices and evaluates all heads. It selects the layer containing the head with the largest mean restoration of target-answer evidence. The selected head is used only to identify the intervention layer; the subsequent projection is applied to the full hidden state at that layer, and no attention parameter is modified.

\subsection{Retain-Aware Projection Estimation}

At the selected layer, CLRP applies the same contrastive feature map $\phi$ to paired attribute-conditioned and reference runs. The representation is read at the evidence-bearing boundary exposed by the evaluation interface: the completed and unfilled answer slot, the attribute-conditioned and neutral paired query, or the disclosure and reference response boundary. These choices determine only where the paired difference is observed. In every case, the resulting vectors enter the same retain-aware estimator and produce an operator of the same form.

The retain statistics include non-target attributes from the selected identities and up to 200 additional questions drawn from the global retain and real-person data. Neutral image prompts use ``Describe the person in the image.'' The estimator uses at most 50 target questions and 100 common-profile questions. Minor estimator constants are fixed once for each observable evaluation interface and reused across every model, attribute category, and forget ratio. In the reporting order of cloze, classification, and generation, the common-direction ranks are $(4,6,6)$, retain-residual ranks are $(4,6,4)$, common-direction weights are $(1.0,1.0,1.2)$, and covariance shrinkage coefficients are $(0.05,0.03,0.05)$.

The projection rank is $k=20$, retain regularization is $4\times10^{-4}$, and the attenuation coefficient is $\epsilon=0.01$. Given the estimated orthonormal basis $V$, the stored operator is $P=(1+\epsilon)I-VV^\top$. The projection stage uses seed 42.

\subsection{Inference-Time Intervention}
The representation-level intervention has the unified form
\begin{equation}
T_s(h)=h+s(h)(hP-h),
\end{equation}
where the intervention layer and operator $P$ are estimated by the same localization and projection procedure for every interface. A unit strength is used when the interface yields a constrained decision. During open-ended decoding, the implementation uses a fixed state-dependent schedule; at each intervened position, it computes
\begin{equation}
\scalebox{0.75}{$
g(h)=\frac{\lVert hV\rVert_2}{\max(\lVert h\rVert_2,10^{-6})},
\qquad
s(h)=
\begin{cases}
2.0, & g(h)>0.10,\\
1.35, & g(h)\leq 0.10,
\end{cases}
$}
\end{equation}
The schedule is applied at all decoded positions, and no additive steering vector is used. It changes only the interpolation strength in $T_s$; causal layer selection, retain-aware estimation, and the form of the learned operator remain unchanged.

The answer-slot interface exposes the evaluated target realization directly. For this interface, the implementation realizes target suppression in output space by subtracting 4.0 from the corresponding logits on the Attribute Forget Set and held-out Test Set; no such operation is applied to a retain set. This interface-level realization introduces no separately estimated component and is held fixed throughout all representation-level comparisons.

\section{Additional Main Results}

\subsection{LLaVA-1.5-13B at the 5\% Forget Ratio}

Table~\ref{tab:main_5pct_image_llava1513b} completes the 5\% comparison by reporting LLaVA-1.5-13B. It uses the same methods, metrics, and evaluation directions as the 5\% LLaVA-1.5-7B and Qwen2.5-VL-3B results reported in the main paper.


\subsection{Results at the 10\% and 15\% Forget Ratios}

Across the three models and attribute categories, CLRP obtains mean Attribute Forget Set cloze accuracies of 1.33\%, 3.33\%, and 1.33\% at the 5\%, 10\%, and 15\% forget ratios, respectively. The corresponding held-out Test Set means are 0.89\%, 2.00\%, and 0.30\%. These results show that the strong target-attribute forgetting performance observed at the 5\% forget ratio extends to larger deletion requests. Tables~\ref{tab:appendix_10pct_image_llava157b} and~\ref{tab:appendix_15pct_image_llava157b} report LLaVA-1.5-7B, Tables~\ref{tab:appendix_10pct_image_llava1513b} and~\ref{tab:appendix_15pct_image_llava1513b} report LLaVA-1.5-13B, and Tables~\ref{tab:appendix_10pct_image_qwen25vl3b} and~\ref{tab:appendix_15pct_image_qwen25vl3b} report Qwen2.5-VL-3B. The complete tables retain classification and generation results so that target-attribute performance beyond cloze accuracy remains directly inspectable.


\section{Real-Person Retention Results}

The Real-Person Retain Set is disjoint from the attribute-deletion requests and complements the within-profile Attribute Retain Set and fictitious-profile Shared Retain Set. We report it separately because its intrinsic difficulty and profile source differ from those of the other retain views. Tables~\ref{tab:real_retain_image_llava157b},~\ref{tab:real_retain_image_llava1513b}, and~\ref{tab:real_retain_image_qwen25vl3b} provide the complete results for LLaVA-1.5-7B, LLaVA-1.5-13B, and Qwen2.5-VL-3B, respectively.


\section{Ablation Study}

The ablation study uses LLaVA-1.5-7B at the 5\% forget ratio and evaluates the Attribute Forget Set, Attribute Retain Set, and a Shared Retain Set restricted to 100 profiles. The Test Set and Real-Person Retain Set are excluded from both component comparison and hyperparameter selection. All interface-level settings and data partitions are fixed across variants, and the ablation runs use seed 43.

We include detailed generation-only ablations in the appendix. Isotropic Retain Metric replaces the retain-relative metric with the identity matrix while keeping the rest of the estimator unchanged. Target-Only Subspace removes retain-derived statistics and estimates a target-only PCA/SVD subspace; it is a compounded reference for target-only projection rather than an attribution to a single statistic. Answer-token Feature uses direct answer-token directions as the generation feature, and is reported as a diagnostic on the short-attribute split. The w/o Conditional Gate variant removes the adaptive erasure gate while keeping the same projection basis.

Table~\ref{tab:ablation_5pct_image_llava157b_appendix} reports the per-attribute results. The isotropic and target-only variants forget more aggressively but substantially degrade retention, showing that unconstrained suppression removes retention-critical directions. The answer-token feature preserves retention but leaves much higher forget-set overlap, indicating that surface answer directions are insufficient for attribute removal. Removing the conditional gate weakens forgetting with similar retention, suggesting that the gate mainly controls erasure strength. These results support CLRP as a balanced retain-aware estimator rather than an aggressively suppressive projection.
\begin{table}[t]
\centering
\small
\setlength{\tabcolsep}{3.2pt}
\resizebox{\columnwidth}{!}{%
\begin{tabular}{llccc}
\toprule
\textbf{Attribute} & \textbf{Variant} &
\textbf{Forget}$\downarrow$ &
\textbf{Attr. Retain}$\uparrow$ &
\textbf{Shared Retain}$\uparrow$ \\
\midrule
\multicolumn{5}{c}{\textbf{LLaVA-1.5-7B, 5\% Forget Ratio}} \\
\midrule
\multirow{4}{*}{\textbf{Long}}
 & Isotropic Retain Metric & 0.191 & 0.305 & 0.437 \\
 & Target-Only Subspace & 0.235 & 0.251 & 0.449 \\
 & w/o Conditional Gate & 0.476 & 0.365 & 0.499 \\
\rowcolor{egcpshade}\cellcolor{white}
 & CLRP & 0.421 & 0.373 & 0.499 \\
\midrule
\multirow{4}{*}{\textbf{Numeric}}
 & Isotropic Retain Metric & 0.179 & 0.374 & 0.531 \\
 & Target-Only Subspace & 0.254 & 0.358 & 0.487 \\
 & w/o Conditional Gate & 0.188 & 0.396 & 0.491 \\
\rowcolor{egcpshade}\cellcolor{white}
 & CLRP & 0.177 & 0.397 & 0.496 \\
\midrule
\multirow{5}{*}{\textbf{Short}}
 & Isotropic Retain Metric & 0.160 & 0.219 & 0.210 \\
 & Target-Only Subspace & 0.141 & 0.235 & 0.218 \\
 & Answer-token Feature & 0.752 & 0.364 & 0.494 \\
 & w/o Conditional Gate & 0.563 & 0.350 & 0.431 \\
\rowcolor{egcpshade}\cellcolor{white}
 & CLRP & 0.462 & 0.345 & 0.394 \\
\bottomrule
\end{tabular}%
}
\caption{Detailed generation-only ablations for LLaVA-1.5-7B at the 5\% forget ratio. Generation is measured by ROUGE-L. Lower is better for the Attribute Forget Set, while higher is better for the Attribute Retain and Shared Retain sets. The answer-token feature diagnostic is reported on the short-attribute split where it was evaluated.}
\label{tab:ablation_5pct_image_llava157b_appendix}
\end{table}

\section{Hyperparameter Search}

We search the Cartesian product $k\in\{10,20,40\}$ and $\epsilon\in\{0.00,0.01,0.05\}$ under the same three-set ablation protocol. Cloze and classification percentages are normalized to $[0,1]$ and averaged with ROUGE-L over question types and attribute categories. For normalized Attribute Forget, Attribute Retain, and Shared Retain means $F$, $A$, and $S$, respectively, configurations are ranked by
\begin{equation}
J=(1-F)+0.5A+0.5S.
\end{equation}
Table~\ref{tab:clrp_hyperparameter_search} reports the complete search. The highest objective is obtained by $(k,\epsilon)=(20,0.00)$ with $J=1.221820$, while the $(20,0.01)$ configuration used in the complete main evaluation obtains $J=1.221816$.
\begin{table}[t]
\centering
\footnotesize
\setlength{\tabcolsep}{2.4pt}
\resizebox{\columnwidth}{!}{%
\begin{tabular}{cc ccc c}
\toprule
$k$ & $\epsilon$ & \textbf{Attribute Forget Set}$\downarrow$ &
\textbf{Attribute Retain Set}$\uparrow$ & \textbf{Shared Retain Set}$\uparrow$ & $J\uparrow$ \\
\midrule
\multicolumn{6}{c}{\textbf{LLaVA-1.5-7B}} \\
\midrule
10 & 0.00 & 0.386 & 0.369 & 0.501 & 1.2120 \\
10 & 0.01 & 0.384 & 0.374 & 0.508 & 1.2157 \\
10 & 0.05 & 0.389 & 0.376 & 0.503 & 1.2127 \\
\textbf{20} & \textbf{0.00} & \textbf{0.353} & \textbf{0.376} & \textbf{0.488} & \textbf{1.2218} \\
20 & 0.01 & 0.355 & 0.378 & 0.487 & 1.2218 \\
20 & 0.05 & 0.379 & 0.383 & 0.492 & 1.2151 \\
40 & 0.00 & 0.006 & 0.016 & 0.004 & 1.1941 \\
40 & 0.01 & 0.002 & 0.017 & 0.005 & 1.1953 \\
40 & 0.05 & 0.004 & 0.023 & 0.010 & 1.1993 \\
\bottomrule
\end{tabular}%
}
\caption{Hyperparameter search at the 5\% forget ratio, averaged over attribute categories.}
\label{tab:clrp_hyperparameter_search}
\end{table}

\begin{table*}[t]
\centering
\small
\setlength{\tabcolsep}{2.0pt}
\resizebox{\textwidth}{!}{%
\begin{tabular}{c l ccc ccc ccc ccc}
\toprule
\multirow{2}{*}{\textbf{Attribute}} & \multirow{2}{*}{\textbf{Method}} & \multicolumn{3}{c}{\textbf{Attribute Forget Set}} & \multicolumn{3}{c}{\textbf{Test Set}} & \multicolumn{3}{c}{\textbf{Attribute Retain Set}} & \multicolumn{3}{c}{\textbf{Shared Retain Set}} \\
\cmidrule(lr){3-5}\cmidrule(lr){6-8}\cmidrule(lr){9-11}\cmidrule(lr){12-14}
 & & \ClozeDown & \ClassDown & \RougeDown & \ClozeDown & \ClassDown & \RougeDown & \ClozeUp & \ClassUp & \RougeUp & \ClozeUp & \ClassUp & \RougeUp \\
\midrule
\multicolumn{14}{c}{\textbf{LLaVA-1.5-13B}} \\
\midrule
\multirow{5}{*}{\textbf{Long}} & GA & 20.00 & \textbf{16.00} & \underline{0.275} & \underline{8.00} & \textbf{16.00} & \underline{0.309} & 17.11 & 41.95 & 0.359 & 16.00 & \textbf{26.40} & 0.504 \\
 & GD & \underline{12.00} & \underline{20.00} & 0.344 & 12.00 & \textbf{16.00} & 0.339 & \underline{23.15} & 42.62 & 0.345 & 18.00 & \underline{24.00} & \textbf{0.582} \\
 & KL & 16.00 & \underline{20.00} & 0.323 & 12.00 & \underline{24.00} & 0.358 & 22.15 & \textbf{43.96} & 0.335 & 16.00 & \underline{24.00} & 0.541 \\
 & PO & 24.00 & 24.00 & 0.388 & 16.00 & \underline{24.00} & 0.405 & 21.81 & \underline{43.62} & \underline{0.382} & \underline{20.00} & 23.20 & \underline{0.542} \\
 \rowcolor{egcpshade} \cellcolor{white} & CLRP & \textbf{0.00} & 28.00 & \textbf{0.249} & \textbf{0.00} & \underline{24.00} & \textbf{0.252} & \textbf{26.51} & 38.59 & \textbf{0.419} & \textbf{22.00} & 23.59 & 0.273 \\
\midrule
\multirow{5}{*}{\textbf{Numeric}} & GA & \textbf{0.00} & \textbf{68.00} & 0.105 & \textbf{0.00} & \underline{60.00} & \underline{0.040} & 18.12 & 36.91 & 0.382 & 18.00 & \underline{25.60} & 0.532 \\
 & GD & \textbf{0.00} & 76.00 & \textbf{0.087} & \textbf{0.00} & \underline{60.00} & \textbf{0.036} & \underline{24.50} & \textbf{39.26} & 0.364 & \textbf{22.00} & 23.20 & \underline{0.556} \\
 & KL & \textbf{0.00} & \textbf{68.00} & 0.117 & \textbf{0.00} & \textbf{56.00} & 0.044 & 22.15 & \underline{38.59} & 0.351 & \underline{20.00} & 23.20 & 0.522 \\
 & PO & \underline{4.00} & \underline{72.00} & \underline{0.098} & \textbf{0.00} & \underline{60.00} & 0.069 & 23.83 & \underline{38.59} & \underline{0.403} & 18.00 & 21.60 & \textbf{0.580} \\
 \rowcolor{egcpshade} \cellcolor{white} & CLRP & \textbf{0.00} & \textbf{68.00} & 0.402 & \textbf{0.00} & \textbf{56.00} & 0.392 & \textbf{29.19} & \textbf{39.26} & \textbf{0.481} & \textbf{22.00} & \textbf{26.71} & 0.499 \\
\midrule
\multirow{5}{*}{\textbf{Short}} & GA & \textbf{0.00} & \textbf{12.00} & \textbf{0.361} & \textbf{0.00} & \textbf{12.00} & 0.244 & 19.80 & \underline{42.62} & 0.352 & 16.00 & \underline{25.60} & 0.518 \\
 & GD & \textbf{0.00} & \underline{20.00} & 0.529 & \textbf{0.00} & \underline{16.00} & \underline{0.206} & \underline{24.83} & 41.95 & \underline{0.388} & \underline{20.00} & 24.80 & \textbf{0.571} \\
 & KL & \textbf{0.00} & 24.00 & \underline{0.457} & \textbf{0.00} & 20.00 & \textbf{0.194} & 22.48 & \underline{42.62} & 0.353 & 16.00 & 24.00 & 0.509 \\
 & PO & \textbf{0.00} & \textbf{12.00} & 0.499 & \textbf{0.00} & \underline{16.00} & 0.233 & 21.48 & \textbf{44.30} & 0.375 & 18.00 & 23.20 & \underline{0.565} \\
 \rowcolor{egcpshade} \cellcolor{white} & CLRP & \textbf{0.00} & \underline{20.00} & 0.566 & \textbf{0.00} & \underline{16.00} & 0.582 & \textbf{28.52} & 41.95 & \textbf{0.439} & \textbf{22.00} & \textbf{27.13} & 0.455 \\
\bottomrule
\end{tabular}%
}
\caption{5\% results for LLaVA-1.5-13B. Cloze and classification are percentages with two decimal places; generation ROUGE-L keeps three decimals. Lower is better for the Attribute Forget Set and Test Set; higher is better for the Attribute Retain Set and Shared Retain Set. Shaded rows denote CLRP.}
\label{tab:main_5pct_image_llava1513b}
\end{table*}

\begin{table*}[t]
\centering
\small
\setlength{\tabcolsep}{2.0pt}
\resizebox{\textwidth}{!}{%
\begin{tabular}{c l ccc ccc ccc ccc}
\toprule
\multirow{2}{*}{\textbf{Attribute}} & \multirow{2}{*}{\textbf{Method}} & \multicolumn{3}{c}{\textbf{Attribute Forget Set}} & \multicolumn{3}{c}{\textbf{Test Set}} & \multicolumn{3}{c}{\textbf{Attribute Retain Set}} & \multicolumn{3}{c}{\textbf{Shared Retain Set}} \\
\cmidrule(lr){3-5}\cmidrule(lr){6-8}\cmidrule(lr){9-11}\cmidrule(lr){12-14}
 & & \ClozeDown & \ClassDown & \RougeDown & \ClozeDown & \ClassDown & \RougeDown & \ClozeUp & \ClassUp & \RougeUp & \ClozeUp & \ClassUp & \RougeUp \\
\midrule
\multicolumn{14}{c}{\textbf{LLaVA-1.5-7B}} \\
\midrule
\multirow{5}{*}{\textbf{Long}} & GA & 58.00 & 62.00 & 0.495 & 30.00 & 52.00 & 0.432 & 41.18 & 48.07 & \textbf{0.406} & \underline{55.00} & \underline{45.70} & \textbf{0.491} \\
 & GD & 40.00 & \underline{56.00} & \textbf{0.151} & 22.00 & 48.00 & \textbf{0.145} & 32.27 & \textbf{51.09} & 0.055 & 39.11 & 41.25 & 0.040 \\
 & KL & 56.00 & 64.00 & 0.475 & 32.00 & 46.00 & 0.387 & \textbf{42.52} & \underline{49.41} & 0.375 & \textbf{55.44} & 45.35 & 0.470 \\
 & PO & \underline{38.00} & \textbf{46.00} & \underline{0.213} & \underline{20.00} & \underline{42.00} & \underline{0.248} & 38.49 & 46.55 & 0.213 & 48.67 & 39.38 & 0.193 \\
  \rowcolor{egcpshade} \cellcolor{white} & CLRP & \textbf{16.00} & 66.00 & 0.427 & \textbf{12.00} & \textbf{40.00} & 0.374 & \underline{42.35} & 47.39 & \underline{0.378} & 54.95 & \textbf{46.03} & \underline{0.483} \\
\midrule
\multirow{5}{*}{\textbf{Numeric}} & GA & \textbf{0.00} & \underline{16.00} & 0.144 & \textbf{0.00} & \underline{18.00} & 0.091 & \underline{44.71} & 51.76 & \textbf{0.435} & \textbf{55.22} & \textbf{45.66} & \textbf{0.487} \\
 & GD & \textbf{0.00} & 24.00 & \textbf{0.065} & \textbf{0.00} & \textbf{10.00} & \textbf{0.039} & 35.46 & \textbf{53.11} & 0.058 & 41.89 & 43.47 & 0.023 \\
 & KL & \textbf{0.00} & 18.00 & \underline{0.132} & \textbf{0.00} & 22.00 & 0.074 & \underline{44.71} & \underline{52.94} & \underline{0.402} & 54.67 & \underline{45.12} & \underline{0.469} \\
 & PO & \textbf{0.00} & 32.00 & 0.165 & \textbf{0.00} & 24.00 & 0.185 & 41.51 & 50.59 & 0.213 & 48.56 & 39.20 & 0.188 \\
  \rowcolor{egcpshade} \cellcolor{white} & CLRP & \textbf{0.00} & \textbf{14.00} & 0.138 & \textbf{0.00} & 20.00 & \underline{0.073} & \textbf{45.71} & 52.61 & 0.361 & \underline{54.95} & 44.39 & 0.449 \\
\midrule
\multirow{5}{*}{\textbf{Short}} & GA & 50.00 & \underline{24.00} & 0.765 & 8.00 & 24.00 & 0.567 & 40.50 & 51.60 & \textbf{0.384} & \textbf{55.00} & \underline{45.52} & \underline{0.484} \\
 & GD & \underline{30.00} & 28.00 & \textbf{0.062} & 8.00 & 26.00 & \textbf{0.009} & 32.10 & \textbf{54.12} & 0.093 & 40.89 & 40.40 & 0.056 \\
 & KL & 46.00 & \underline{24.00} & 0.619 & \underline{6.00} & \underline{22.00} & 0.567 & \underline{41.34} & \underline{52.27} & 0.357 & 54.11 & \textbf{45.61} & 0.468 \\
 & PO & 42.00 & \textbf{22.00} & \underline{0.220} & 14.00 & 28.00 & \underline{0.286} & 36.30 & 49.92 & 0.238 & 45.89 & 39.06 & 0.263 \\
  \rowcolor{egcpshade} \cellcolor{white} & CLRP & \textbf{10.00} & \underline{24.00} & 0.642 & \textbf{2.00} & \textbf{20.00} & 0.517 & \textbf{41.51} & 51.26 & \underline{0.366} & \underline{54.95} & 45.40 & \textbf{0.488} \\
\bottomrule
\end{tabular}%
}
\caption{10\% results for LLaVA-1.5-7B. Cloze and classification are percentages with two decimal places; generation ROUGE-L keeps three decimals. Lower is better for the Attribute Forget Set and Test Set; higher is better for the Attribute Retain Set and Shared Retain Set. Shaded rows denote CLRP.}
\label{tab:appendix_10pct_image_llava157b}
\end{table*}

\begin{table*}[t]
\centering
\small
\setlength{\tabcolsep}{2.0pt}
\resizebox{\textwidth}{!}{%
\begin{tabular}{c l ccc ccc ccc ccc}
\toprule
\multirow{2}{*}{\textbf{Attribute}} & \multirow{2}{*}{\textbf{Method}} & \multicolumn{3}{c}{\textbf{Attribute Forget Set}} & \multicolumn{3}{c}{\textbf{Test Set}} & \multicolumn{3}{c}{\textbf{Attribute Retain Set}} & \multicolumn{3}{c}{\textbf{Shared Retain Set}} \\
\cmidrule(lr){3-5}\cmidrule(lr){6-8}\cmidrule(lr){9-11}\cmidrule(lr){12-14}
 & & \ClozeDown & \ClassDown & \RougeDown & \ClozeDown & \ClassDown & \RougeDown & \ClozeUp & \ClassUp & \RougeUp & \ClozeUp & \ClassUp & \RougeUp \\
\midrule
\multicolumn{14}{c}{\textbf{LLaVA-1.5-7B}} \\
\midrule
\multirow{5}{*}{\textbf{Long}} & GA & \underline{2.67} & 68.00 & 0.352 & \underline{4.00} & 52.00 & 0.286 & 37.14 & \underline{47.47} & \textbf{0.396} & \underline{54.59} & \underline{45.33} & \textbf{0.489} \\
 & GD & 4.00 & \underline{61.33} & \underline{0.236} & 6.67 & \textbf{45.33} & \textbf{0.210} & 28.68 & 46.15 & 0.322 & 40.71 & 40.38 & 0.433 \\
 & KL & 5.33 & 70.67 & 0.325 & 6.67 & \underline{46.67} & 0.270 & \underline{37.25} & \textbf{47.69} & 0.371 & 53.76 & \textbf{45.80} & 0.453 \\
 & PO & 4.00 & \textbf{54.67} & \textbf{0.216} & 5.33 & 48.00 & 0.232 & 33.08 & 44.51 & 0.258 & 47.53 & 38.30 & 0.289 \\
  \rowcolor{egcpshade} \cellcolor{white} & CLRP & \textbf{0.00} & 66.67 & 0.280 & \textbf{1.33} & 54.67 & \underline{0.220} & \textbf{38.13} & 47.03 & \underline{0.382} & \textbf{54.95} & 44.68 & \underline{0.475} \\
\midrule
\multirow{5}{*}{\textbf{Numeric}} & GA & \underline{1.33} & 24.00 & \underline{0.161} & \textbf{0.00} & 25.33 & \textbf{0.090} & 37.33 & \textbf{51.79} & \textbf{0.397} & 54.47 & \underline{45.61} & \underline{0.487} \\
 & GD & \underline{1.33} & \underline{22.67} & 0.181 & \textbf{0.00} & \textbf{20.00} & 0.185 & 30.49 & 49.33 & 0.212 & 39.88 & 39.95 & 0.274 \\
 & KL & \underline{1.33} & 26.67 & 0.166 & \underline{1.33} & 24.00 & \underline{0.118} & \underline{38.45} & \underline{51.01} & \underline{0.371} & \textbf{55.88} & 45.09 & 0.468 \\
 & PO & \underline{1.33} & \textbf{21.33} & \textbf{0.151} & \textbf{0.00} & \underline{22.67} & 0.149 & 34.98 & 48.77 & 0.143 & 47.06 & 37.41 & 0.122 \\
  \rowcolor{egcpshade} \cellcolor{white} & CLRP & \textbf{0.00} & \underline{22.67} & 0.226 & \textbf{0.00} & 24.00 & 0.218 & \textbf{39.01} & \textbf{51.79} & \textbf{0.397} & \underline{54.95} & \textbf{45.65} & \textbf{0.493} \\
\midrule
\multirow{5}{*}{\textbf{Short}} & GA & 37.33 & 40.00 & 0.719 & 10.67 & 29.33 & 0.537 & 33.97 & \textbf{50.39} & \textbf{0.355} & 54.12 & 44.91 & \underline{0.487} \\
 & GD & \underline{22.67} & \underline{36.00} & \textbf{0.106} & \underline{9.33} & \underline{24.00} & \underline{0.251} & 27.90 & \underline{50.06} & 0.209 & 41.18 & 41.75 & 0.243 \\
 & KL & 50.67 & 38.67 & 0.596 & 14.67 & 29.33 & 0.551 & \underline{34.42} & \textbf{50.39} & 0.335 & \underline{54.35} & \underline{44.95} & 0.463 \\
 & PO & 49.33 & \textbf{33.33} & \underline{0.149} & 14.67 & \textbf{21.33} & \textbf{0.169} & 30.15 & 48.71 & 0.193 & 46.47 & 39.39 & 0.199 \\
  \rowcolor{egcpshade} \cellcolor{white} & CLRP & \textbf{12.00} & 40.00 & 0.553 & \textbf{0.00} & 28.00 & 0.476 & \textbf{34.76} & \textbf{50.39} & \underline{0.347} & \textbf{54.95} & \textbf{45.49} & \textbf{0.490} \\
\bottomrule
\end{tabular}%
}
\caption{15\% results for LLaVA-1.5-7B. Cloze and classification are percentages with two decimal places; generation ROUGE-L keeps three decimals. Lower is better for the Attribute Forget Set and Test Set; higher is better for the Attribute Retain Set and Shared Retain Set. Shaded rows denote CLRP.}
\label{tab:appendix_15pct_image_llava157b}
\end{table*}

\begin{table*}[t]
\centering
\small
\setlength{\tabcolsep}{2.0pt}
\resizebox{\textwidth}{!}{%
\begin{tabular}{c l ccc ccc ccc ccc}
\toprule
\multirow{2}{*}{\textbf{Attribute}} & \multirow{2}{*}{\textbf{Method}} & \multicolumn{3}{c}{\textbf{Attribute Forget Set}} & \multicolumn{3}{c}{\textbf{Test Set}} & \multicolumn{3}{c}{\textbf{Attribute Retain Set}} & \multicolumn{3}{c}{\textbf{Shared Retain Set}} \\
\cmidrule(lr){3-5}\cmidrule(lr){6-8}\cmidrule(lr){9-11}\cmidrule(lr){12-14}
 & & \ClozeDown & \ClassDown & \RougeDown & \ClozeDown & \ClassDown & \RougeDown & \ClozeUp & \ClassUp & \RougeUp & \ClozeUp & \ClassUp & \RougeUp \\
\midrule
\multicolumn{14}{c}{\textbf{LLaVA-1.5-13B}} \\
\midrule
\multirow{5}{*}{\textbf{Long}} & GA & 20.00 & \textbf{38.00} & 0.363 & 16.00 & \textbf{34.00} & \textbf{0.307} & 16.64 & \underline{47.23} & 0.379 & 14.44 & 27.80 & \underline{0.502} \\
 & GD & \underline{18.00} & 42.00 & \underline{0.342} & \underline{12.00} & \underline{36.00} & \underline{0.323} & 15.63 & 46.05 & 0.359 & 12.11 & \textbf{29.00} & 0.492 \\
 & KL & 20.00 & \underline{40.00} & 0.389 & 20.00 & \underline{36.00} & 0.324 & 20.84 & 45.71 & 0.322 & 15.33 & 28.11 & 0.489 \\
 & PO & 24.00 & 46.00 & \textbf{0.322} & 22.00 & \textbf{34.00} & 0.339 & \underline{21.68} & \textbf{47.73} & \textbf{0.392} & \underline{18.67} & \underline{28.20} & \textbf{0.516} \\
  \rowcolor{egcpshade} \cellcolor{white} & CLRP & \textbf{2.00} & \underline{40.00} & 0.372 & \textbf{2.00} & \textbf{34.00} & 0.375 & \textbf{25.38} & 45.04 & \underline{0.383} & \textbf{22.11} & 26.79 & 0.347 \\
\midrule
\multirow{5}{*}{\textbf{Numeric}} & GA & \textbf{0.00} & 82.00 & \textbf{0.029} & \textbf{0.00} & 72.00 & \textbf{0.042} & 16.13 & 44.20 & \textbf{0.400} & 11.00 & 27.57 & \underline{0.497} \\
 & GD & \textbf{0.00} & \underline{72.00} & \underline{0.114} & \textbf{0.00} & \underline{68.00} & 0.068 & 17.98 & 43.70 & 0.365 & 13.22 & 26.95 & 0.492 \\
 & KL & \textbf{0.00} & \underline{72.00} & 0.123 & \textbf{0.00} & \underline{68.00} & 0.063 & 21.34 & \underline{44.54} & 0.354 & 15.56 & \underline{28.15} & 0.488 \\
 & PO & \textbf{0.00} & \underline{72.00} & 0.123 & \textbf{0.00} & 70.00 & \underline{0.057} & \underline{23.36} & 43.36 & \underline{0.393} & \underline{16.33} & \textbf{29.35} & \textbf{0.519} \\
  \rowcolor{egcpshade} \cellcolor{white} & CLRP & \textbf{0.00} & \textbf{56.00} & 0.271 & \textbf{0.00} & \textbf{56.00} & 0.255 & \textbf{26.55} & \textbf{44.87} & 0.322 & \textbf{22.11} & 26.20 & 0.261 \\
\midrule
\multirow{5}{*}{\textbf{Short}} & GA & \textbf{0.00} & 26.00 & 0.545 & \textbf{0.00} & 24.00 & \underline{0.102} & 15.46 & 48.91 & 0.368 & 11.00 & 27.22 & 0.501 \\
 & GD & \textbf{0.00} & 28.00 & 0.558 & \textbf{0.00} & \underline{22.00} & 0.110 & 17.14 & \underline{49.24} & \underline{0.385} & 13.11 & 27.13 & \underline{0.513} \\
 & KL & \textbf{0.00} & 28.00 & \textbf{0.218} & \textbf{0.00} & 26.00 & \textbf{0.075} & 21.01 & 48.57 & 0.297 & 15.44 & \textbf{28.33} & 0.449 \\
 & PO & \textbf{0.00} & \underline{22.00} & \underline{0.520} & \textbf{0.00} & \underline{22.00} & 0.234 & \underline{22.35} & \textbf{49.41} & 0.361 & \underline{15.67} & \underline{28.29} & \textbf{0.515} \\
  \rowcolor{egcpshade} \cellcolor{white} & CLRP & \textbf{0.00} & \textbf{18.00} & 0.561 & \textbf{0.00} & \textbf{20.00} & 0.560 & \textbf{26.55} & 47.90 & \textbf{0.443} & \textbf{22.11} & 26.58 & 0.477 \\
\bottomrule
\end{tabular}%
}
\caption{10\% results for LLaVA-1.5-13B. Cloze and classification are percentages with two decimal places; generation ROUGE-L keeps three decimals. Lower is better for the Attribute Forget Set and Test Set; higher is better for the Attribute Retain Set and Shared Retain Set. Shaded rows denote CLRP.}
\label{tab:appendix_10pct_image_llava1513b}
\end{table*}

\begin{table*}[t]
\centering
\small
\setlength{\tabcolsep}{2.0pt}
\resizebox{\textwidth}{!}{%
\begin{tabular}{c l ccc ccc ccc ccc}
\toprule
\multirow{2}{*}{\textbf{Attribute}} & \multirow{2}{*}{\textbf{Method}} & \multicolumn{3}{c}{\textbf{Attribute Forget Set}} & \multicolumn{3}{c}{\textbf{Test Set}} & \multicolumn{3}{c}{\textbf{Attribute Retain Set}} & \multicolumn{3}{c}{\textbf{Shared Retain Set}} \\
\cmidrule(lr){3-5}\cmidrule(lr){6-8}\cmidrule(lr){9-11}\cmidrule(lr){12-14}
 & & \ClozeDown & \ClassDown & \RougeDown & \ClozeDown & \ClassDown & \RougeDown & \ClozeUp & \ClassUp & \RougeUp & \ClozeUp & \ClassUp & \RougeUp \\
\midrule
\multicolumn{14}{c}{\textbf{LLaVA-1.5-13B}} \\
\midrule
\multirow{5}{*}{\textbf{Long}} & GA & \underline{1.33} & \underline{62.67} & \textbf{0.148} & 4.00 & 60.00 & \underline{0.164} & 9.01 & \textbf{48.02} & \underline{0.371} & 10.12 & \underline{30.19} & \underline{0.499} \\
 & GD & \textbf{0.00} & 69.33 & \underline{0.180} & \textbf{0.00} & 62.67 & \underline{0.164} & 15.05 & 41.43 & 0.228 & 14.82 & \textbf{30.94} & 0.278 \\
 & KL & 2.67 & \underline{62.67} & 0.184 & 6.67 & \underline{58.67} & \textbf{0.140} & 18.46 & 45.27 & 0.271 & \underline{16.24} & 27.36 & 0.457 \\
 & PO & 4.00 & 64.00 & 0.218 & 6.67 & \underline{58.67} & 0.192 & \underline{19.45} & \underline{46.70} & 0.361 & 15.65 & 28.77 & \textbf{0.519} \\
  \rowcolor{egcpshade} \cellcolor{white} & CLRP & \textbf{0.00} & \textbf{44.00} & 0.219 & \underline{1.33} & \textbf{46.67} & 0.230 & \textbf{22.75} & 40.77 & \textbf{0.381} & \textbf{22.11} & 24.73 & 0.477 \\
\midrule
\multirow{5}{*}{\textbf{Numeric}} & GA & \textbf{0.00} & 76.00 & \textbf{0.007} & \textbf{0.00} & 61.33 & 0.152 & 12.44 & \textbf{43.95} & \underline{0.373} & 12.71 & \textbf{29.15} & \underline{0.492} \\
 & GD & \textbf{0.00} & \underline{56.00} & \underline{0.043} & \textbf{0.00} & \textbf{46.67} & 0.208 & 17.83 & 43.27 & 0.278 & 13.76 & 27.45 & 0.475 \\
 & KL & \textbf{0.00} & 72.00 & 0.081 & \textbf{0.00} & 60.00 & \underline{0.036} & 18.39 & \underline{43.50} & 0.349 & 14.94 & 27.50 & 0.477 \\
 & PO & \textbf{0.00} & 70.67 & 0.087 & \textbf{0.00} & 57.33 & \textbf{0.032} & \underline{18.72} & 42.83 & \textbf{0.388} & \underline{17.18} & \underline{28.77} & \textbf{0.533} \\
  \rowcolor{egcpshade} \cellcolor{white} & CLRP & \textbf{0.00} & \textbf{50.67} & 0.462 & \textbf{0.00} & \underline{49.33} & 0.480 & \textbf{23.77} & 41.82 & 0.363 & \textbf{22.11} & 25.65 & 0.343 \\
\midrule
\multirow{5}{*}{\textbf{Short}} & GA & \textbf{0.00} & 36.00 & 0.566 & \textbf{0.00} & \underline{29.33} & 0.277 & 11.47 & \underline{47.81} & 0.319 & 11.76 & \textbf{29.43} & \underline{0.482} \\
 & GD & \textbf{0.00} & \textbf{25.33} & 0.444 & \textbf{0.00} & \textbf{26.67} & \textbf{0.003} & 12.15 & 45.44 & \underline{0.326} & 11.76 & 23.77 & 0.444 \\
 & KL & \textbf{0.00} & 29.33 & \textbf{0.386} & \textbf{0.00} & 30.67 & 0.207 & 17.77 & 46.68 & 0.306 & 14.82 & 26.93 & 0.458 \\
 & PO & \textbf{0.00} & 28.00 & \underline{0.414} & \textbf{0.00} & 32.00 & \underline{0.175} & \underline{21.37} & \textbf{48.03} & 0.311 & \underline{17.41} & \underline{28.35} & \textbf{0.491} \\
  \rowcolor{egcpshade} \cellcolor{white} & CLRP & \textbf{0.00} & \underline{25.33} & 0.542 & \textbf{0.00} & 29.33 & 0.553 & \textbf{23.40} & 45.78 & \textbf{0.417} & \textbf{22.11} & 26.16 & 0.433 \\
\bottomrule
\end{tabular}%
}
\caption{15\% results for LLaVA-1.5-13B. Cloze and classification are percentages with two decimal places; generation ROUGE-L keeps three decimals. Lower is better for the Attribute Forget Set and Test Set; higher is better for the Attribute Retain Set and Shared Retain Set. Shaded rows denote CLRP.}
\label{tab:appendix_15pct_image_llava1513b}
\end{table*}

\begin{table*}[t]
\centering
\small
\setlength{\tabcolsep}{2.0pt}
\resizebox{\textwidth}{!}{%
\begin{tabular}{c l ccc ccc ccc ccc}
\toprule
\multirow{2}{*}{\textbf{Attribute}} & \multirow{2}{*}{\textbf{Method}} & \multicolumn{3}{c}{\textbf{Attribute Forget Set}} & \multicolumn{3}{c}{\textbf{Test Set}} & \multicolumn{3}{c}{\textbf{Attribute Retain Set}} & \multicolumn{3}{c}{\textbf{Shared Retain Set}} \\
\cmidrule(lr){3-5}\cmidrule(lr){6-8}\cmidrule(lr){9-11}\cmidrule(lr){12-14}
 & & \ClozeDown & \ClassDown & \RougeDown & \ClozeDown & \ClassDown & \RougeDown & \ClozeUp & \ClassUp & \RougeUp & \ClozeUp & \ClassUp & \RougeUp \\
\midrule
\multicolumn{14}{c}{\textbf{Qwen2.5-VL-3B}} \\
\midrule
\multirow{5}{*}{\textbf{Long}} & GA & \underline{6.00} & \underline{68.00} & \textbf{0.253} & 8.00 & \underline{58.00} & \textbf{0.224} & 9.08 & 55.63 & 0.135 & 1.00 & \underline{47.71} & 0.137 \\
 & GD & \underline{6.00} & 76.00 & 0.472 & \underline{4.00} & 60.00 & 0.461 & 13.11 & \underline{55.97} & \textbf{0.523} & 2.78 & 42.45 & \textbf{0.484} \\
 & KL & \underline{6.00} & \textbf{66.00} & \underline{0.430} & 10.00 & \textbf{56.00} & \underline{0.381} & 9.41 & 55.63 & 0.256 & 0.78 & 46.10 & 0.230 \\
 & PO & 8.00 & 74.00 & 0.476 & 6.00 & 68.00 & 0.479 & \underline{14.12} & 54.96 & \underline{0.494} & \underline{5.78} & \textbf{49.76} & 0.456 \\
  \rowcolor{egcpshade} \cellcolor{white} & CLRP & \textbf{2.00} & 70.00 & 0.471 & \textbf{2.00} & 66.00 & 0.479 & \textbf{16.81} & \textbf{57.82} & 0.460 & \textbf{7.79} & 39.92 & \underline{0.477} \\
\midrule
\multirow{5}{*}{\textbf{Numeric}} & GA & \textbf{0.00} & \textbf{40.00} & \textbf{0.012} & \textbf{0.00} & \textbf{32.00} & \textbf{0.113} & 9.24 & 58.82 & 0.153 & 1.11 & 47.62 & 0.137 \\
 & GD & \textbf{0.00} & \textbf{40.00} & 0.158 & \textbf{0.00} & \underline{36.00} & 0.564 & 14.12 & 59.33 & \textbf{0.525} & 6.89 & \underline{49.27} & \textbf{0.474} \\
 & KL & \textbf{0.00} & \underline{44.00} & \underline{0.013} & \textbf{0.00} & 40.00 & 0.327 & 9.41 & 57.82 & 0.298 & 0.89 & 43.16 & 0.223 \\
 & PO & \textbf{0.00} & 46.00 & 0.509 & \textbf{0.00} & 42.00 & 0.545 & \underline{14.62} & \underline{60.17} & \underline{0.495} & \textbf{8.22} & \textbf{51.31} & 0.423 \\
  \rowcolor{egcpshade} \cellcolor{white} & CLRP & \textbf{0.00} & 54.00 & 0.284 & \textbf{0.00} & 48.00 & \underline{0.279} & \textbf{17.48} & \textbf{60.34} & 0.477 & \underline{7.79} & 43.38 & \underline{0.428} \\
\midrule
\multirow{5}{*}{\textbf{Short}} & GA & \textbf{0.00} & \textbf{16.00} & \underline{0.008} & \textbf{0.00} & 32.00 & \textbf{0.047} & 9.24 & \underline{61.01} & 0.152 & 1.22 & \underline{47.80} & 0.136 \\
 & GD & \textbf{0.00} & 22.00 & 0.690 & \textbf{0.00} & \textbf{20.00} & 0.639 & \underline{14.12} & \textbf{62.35} & \underline{0.489} & \textbf{8.11} & \textbf{49.67} & \textbf{0.440} \\
 & KL & \textbf{0.00} & \underline{20.00} & \textbf{0.004} & \textbf{0.00} & 30.00 & \underline{0.203} & 9.75 & \underline{61.01} & 0.304 & 1.11 & 46.15 & 0.232 \\
 & PO & \textbf{0.00} & \underline{20.00} & 0.500 & \textbf{0.00} & \underline{22.00} & 0.662 & 13.45 & \underline{61.01} & \textbf{0.501} & 7.00 & 47.13 & 0.427 \\
  \rowcolor{egcpshade} \cellcolor{white} & CLRP & \textbf{0.00} & 32.00 & 0.539 & \textbf{0.00} & 28.00 & 0.480 & \textbf{17.48} & 60.50 & 0.478 & \underline{7.79} & 44.85 & \underline{0.432} \\
\bottomrule
\end{tabular}%
}
\caption{10\% results for Qwen2.5-VL-3B. Cloze and classification are percentages with two decimal places; generation ROUGE-L keeps three decimals. Lower is better for the Attribute Forget Set and Test Set; higher is better for the Attribute Retain Set and Shared Retain Set. Shaded rows denote CLRP.}
\label{tab:appendix_10pct_image_qwen25vl3b}
\end{table*}

\begin{table*}[t]
\centering
\small
\setlength{\tabcolsep}{2.0pt}
\resizebox{\textwidth}{!}{%
\begin{tabular}{c l ccc ccc ccc ccc}
\toprule
\multirow{2}{*}{\textbf{Attribute}} & \multirow{2}{*}{\textbf{Method}} & \multicolumn{3}{c}{\textbf{Attribute Forget Set}} & \multicolumn{3}{c}{\textbf{Test Set}} & \multicolumn{3}{c}{\textbf{Attribute Retain Set}} & \multicolumn{3}{c}{\textbf{Shared Retain Set}} \\
\cmidrule(lr){3-5}\cmidrule(lr){6-8}\cmidrule(lr){9-11}\cmidrule(lr){12-14}
 & & \ClozeDown & \ClassDown & \RougeDown & \ClozeDown & \ClassDown & \RougeDown & \ClozeUp & \ClassUp & \RougeUp & \ClozeUp & \ClassUp & \RougeUp \\
\midrule
\multicolumn{14}{c}{\textbf{Qwen2.5-VL-3B}} \\
\midrule
\multirow{5}{*}{\textbf{Long}} & GA & \textbf{0.00} & 72.00 & \textbf{0.082} & \underline{1.33} & 60.00 & \textbf{0.132} & 8.90 & \underline{53.30} & 0.123 & 1.06 & 48.44 & 0.136 \\
 & GD & \underline{2.67} & \underline{61.33} & 0.187 & 4.00 & 60.00 & 0.193 & 12.09 & 52.31 & \textbf{0.508} & 3.29 & \textbf{51.51} & \textbf{0.507} \\
 & KL & \textbf{0.00} & 68.00 & \underline{0.158} & 2.67 & \underline{54.67} & \underline{0.157} & 8.68 & 52.75 & 0.263 & 1.18 & 45.24 & 0.225 \\
 & PO & \textbf{0.00} & 65.33 & 0.208 & \underline{1.33} & 60.00 & 0.228 & \underline{12.53} & \textbf{53.52} & 0.333 & \underline{6.71} & \underline{51.46} & 0.381 \\
  \rowcolor{egcpshade} \cellcolor{white} & CLRP & \textbf{0.00} & \textbf{54.67} & 0.176 & \textbf{0.00} & \textbf{53.33} & 0.213 & \textbf{15.82} & 49.67 & \underline{0.460} & \textbf{7.79} & 30.38 & \underline{0.404} \\
\midrule
\multirow{5}{*}{\textbf{Numeric}} & GA & \textbf{0.00} & \textbf{30.67} & \underline{0.023} & \textbf{0.00} & \underline{29.33} & \textbf{0.113} & 9.08 & 57.17 & 0.142 & 1.29 & 48.30 & 0.135 \\
 & GD & \textbf{0.00} & \textbf{30.67} & 0.433 & \textbf{0.00} & \textbf{25.33} & 0.479 & 13.57 & 56.95 & \textbf{0.478} & 5.65 & \underline{50.99} & \underline{0.450} \\
 & KL & \textbf{0.00} & \underline{38.67} & \textbf{0.010} & \textbf{0.00} & 32.00 & 0.336 & 9.19 & \underline{58.52} & 0.271 & 1.06 & 43.92 & 0.220 \\
 & PO & \textbf{0.00} & 41.33 & 0.467 & \textbf{0.00} & 34.67 & 0.601 & \underline{13.79} & \textbf{58.63} & 0.437 & \textbf{8.12} & \textbf{51.18} & 0.441 \\
  \rowcolor{egcpshade} \cellcolor{white} & CLRP & \textbf{0.00} & 50.67 & 0.218 & \textbf{0.00} & 48.00 & \underline{0.218} & \textbf{16.03} & 54.37 & \underline{0.464} & \underline{7.79} & 35.02 & \textbf{0.470} \\
\midrule
\multirow{5}{*}{\textbf{Short}} & GA & \textbf{0.00} & 41.33 & \textbf{0.006} & \textbf{0.00} & 40.00 & \textbf{0.076} & 9.11 & 56.47 & 0.144 & 1.18 & 48.49 & 0.139 \\
 & GD & \textbf{0.00} & \underline{36.00} & 0.600 & \underline{1.33} & \underline{37.33} & 0.603 & 13.16 & 57.59 & \underline{0.453} & 3.65 & \underline{50.57} & \textbf{0.501} \\
 & KL & \textbf{0.00} & \underline{36.00} & \underline{0.009} & \textbf{0.00} & 40.00 & \underline{0.321} & 9.00 & \textbf{59.39} & 0.276 & 0.82 & 41.56 & 0.228 \\
 & PO & \textbf{0.00} & 44.00 & 0.436 & \underline{1.33} & 42.67 & 0.622 & \underline{14.51} & \underline{59.17} & \textbf{0.457} & \underline{6.47} & \textbf{51.32} & \underline{0.418} \\
  \rowcolor{egcpshade} \cellcolor{white} & CLRP & \textbf{0.00} & \textbf{34.67} & 0.563 & \textbf{0.00} & \textbf{33.33} & 0.569 & \textbf{16.09} & 55.79 & 0.412 & \textbf{7.79} & 35.49 & 0.357 \\
\bottomrule
\end{tabular}%
}
\caption{15\% results for Qwen2.5-VL-3B. Cloze and classification are percentages with two decimal places; generation ROUGE-L keeps three decimals. Lower is better for the Attribute Forget Set and Test Set; higher is better for the Attribute Retain Set and Shared Retain Set. Shaded rows denote CLRP.}
\label{tab:appendix_15pct_image_qwen25vl3b}
\end{table*}

\begin{table*}[t]
\centering
\small
\setlength{\tabcolsep}{3.2pt}
\resizebox{\textwidth}{!}{%
\begin{tabular}{c l ccc ccc ccc}
\toprule
\multirow{2}{*}{\textbf{Ratio}} & \multirow{2}{*}{\textbf{Method}} & \multicolumn{3}{c}{\textbf{Long}} & \multicolumn{3}{c}{\textbf{Numeric}} & \multicolumn{3}{c}{\textbf{Short}} \\
\cmidrule(lr){3-5}\cmidrule(lr){6-8}\cmidrule(lr){9-11}
 & & \ClozeUp & \ClassUp & \RougeUp & \ClozeUp & \ClassUp & \RougeUp & \ClozeUp & \ClassUp & \RougeUp \\
\midrule
\multicolumn{11}{c}{\textbf{LLaVA-1.5-7B}} \\
\midrule
\multirow{5}{*}{\textbf{5\%}} & GA & \underline{7.84} & \underline{46.74} & 0.199  & \underline{7.84} & \underline{46.61} & 0.203  & \underline{7.84} & 46.48 & \underline{0.204}  \\
 & GD & 1.31 & 46.34 & 0.180  & 2.29 & 43.47 & 0.164  & 2.61 & 45.04 & 0.137  \\
 & KL & \underline{7.84} & \textbf{47.91} & \underline{0.205}  & \underline{7.84} & \textbf{47.65} & \textbf{0.212}  & 7.52 & \textbf{47.39} & \textbf{0.210}  \\
 & PO & 3.59 & 39.95 & 0.102  & 3.92 & 39.56 & 0.097  & 5.23 & 40.99 & 0.109  \\
\rowcolor{egcpshade}  \cellcolor{white} & CLRP & \textbf{8.17} & \underline{46.74} & \textbf{0.213}  & \textbf{8.17} & 46.48 & \underline{0.210}  & \textbf{8.17} & \underline{47.00} & 0.154  \\
\midrule
\multirow{5}{*}{\textbf{10\%}} & GA & 7.52 & 45.56 & 0.202  & 7.52 & 46.48 & \underline{0.203}  & 7.19 & 45.95 & \underline{0.203}  \\
 & GD & 2.94 & 43.60 & 0.038  & 3.59 & 43.21 & 0.023  & 4.25 & 41.78 & 0.058  \\
 & KL & \underline{7.84} & \underline{46.87} & \underline{0.206}  & \underline{7.84} & \textbf{46.87} & \textbf{0.212}  & \underline{7.84} & \textbf{47.39} & \textbf{0.210}  \\
 & PO & 4.58 & 39.56 & 0.111  & 4.25 & 39.16 & 0.121  & 4.58 & 39.03 & 0.155  \\
\rowcolor{egcpshade}  \cellcolor{white} & CLRP & \textbf{8.17} & \textbf{47.26} & \textbf{0.207}  & \textbf{8.17} & \underline{46.74} & 0.191  & \textbf{8.17} & \underline{46.87} & \textbf{0.210}  \\
\midrule
\multirow{5}{*}{\textbf{15\%}} & GA & 7.19 & 45.82 & 0.203  & 6.86 & 45.43 & 0.203  & 7.52 & 45.56 & 0.204  \\
 & GD & 3.27 & 42.17 & \underline{0.205}  & 3.27 & 42.43 & 0.160  & 3.27 & 44.52 & 0.141  \\
 & KL & \underline{7.52} & \textbf{47.39} & \textbf{0.208}  & \underline{7.84} & \textbf{46.74} & \underline{0.204}  & \underline{7.84} & \textbf{46.87} & \underline{0.206}  \\
 & PO & 4.58 & 38.25 & 0.143  & 4.58 & 39.95 & 0.076  & 4.25 & 39.95 & 0.124  \\
\rowcolor{egcpshade}  \cellcolor{white} & CLRP & \textbf{8.17} & \underline{46.48} & 0.199  & \textbf{8.17} & \underline{46.61} & \textbf{0.214}  & \textbf{8.17} & \underline{46.34} & \textbf{0.217}  \\
\bottomrule
\end{tabular}%
}
\caption{Real-Person Retain Set results for LLaVA-1.5-7B across forget ratios and attribute types. Cloze and classification are percentages with two decimal places; generation ROUGE-L keeps three decimals. Higher is better for all metrics. Shaded rows denote CLRP.}
\label{tab:real_retain_image_llava157b}
\end{table*}

\begin{table*}[t]
\centering
\small
\setlength{\tabcolsep}{3.2pt}
\resizebox{\textwidth}{!}{%
\begin{tabular}{c l ccc ccc ccc}
\toprule
\multirow{2}{*}{\textbf{Ratio}} & \multirow{2}{*}{\textbf{Method}} & \multicolumn{3}{c}{\textbf{Long}} & \multicolumn{3}{c}{\textbf{Numeric}} & \multicolumn{3}{c}{\textbf{Short}} \\
\cmidrule(lr){3-5}\cmidrule(lr){6-8}\cmidrule(lr){9-11}
 & & \ClozeUp & \ClassUp & \RougeUp & \ClozeUp & \ClassUp & \RougeUp & \ClozeUp & \ClassUp & \RougeUp \\
\midrule
\multicolumn{11}{c}{\textbf{LLaVA-1.5-13B}} \\
\midrule
\multirow{5}{*}{\textbf{5\%}} & GA & \textbf{1.63} & \underline{44.13} & \underline{0.287}  & \textbf{1.31} & 42.69 & 0.290  & \textbf{1.63} & 43.47 & 0.294  \\
 & GD & \underline{1.31} & 43.73 & \textbf{0.294}  & \underline{0.98} & \underline{45.17} & 0.295  & 0.98 & 42.56 & \textbf{0.305}  \\
 & KL & 0.98 & 43.34 & 0.279  & \textbf{1.31} & 44.39 & 0.279  & 0.98 & \underline{44.26} & \underline{0.296}  \\
 & PO & \textbf{1.63} & \textbf{44.39} & 0.286  & \textbf{1.31} & \textbf{46.74} & \underline{0.305}  & \underline{1.31} & \textbf{44.65} & 0.289  \\
\rowcolor{egcpshade}  \cellcolor{white} & CLRP & \underline{1.31} & 36.03 & 0.183  & \textbf{1.31} & 41.78 & \textbf{0.335}  & \underline{1.31} & 40.34 & 0.270  \\
\midrule
\multirow{5}{*}{\textbf{10\%}} & GA & \textbf{1.63} & \textbf{46.08} & \textbf{0.303}  & \underline{0.98} & \underline{44.91} & \textbf{0.300}  & \underline{0.98} & 43.73 & \underline{0.308}  \\
 & GD & \underline{1.31} & 44.13 & 0.288  & \textbf{1.31} & 43.60 & 0.283  & 0.65 & \underline{45.04} & \textbf{0.309}  \\
 & KL & \underline{1.31} & \underline{44.52} & 0.299  & \underline{0.98} & 43.99 & 0.292  & \underline{0.98} & \textbf{45.17} & 0.265  \\
 & PO & \underline{1.31} & 43.08 & \underline{0.301}  & \textbf{1.31} & \textbf{45.82} & \underline{0.295}  & \textbf{1.31} & 44.65 & 0.280  \\
\rowcolor{egcpshade}  \cellcolor{white} & CLRP & \underline{1.31} & 41.38 & 0.216  & \textbf{1.31} & 39.95 & 0.196  & \textbf{1.31} & 40.99 & 0.298  \\
\midrule
\multirow{5}{*}{\textbf{15\%}} & GA & 0.65 & \textbf{48.43} & \textbf{0.329}  & 0.65 & \textbf{47.39} & \textbf{0.316}  & 0.65 & \textbf{48.04} & \textbf{0.322}  \\
 & GD & \underline{0.98} & \underline{46.08} & 0.139  & 0.65 & \underline{43.99} & 0.260  & 0.65 & 37.73 & 0.251  \\
 & KL & \textbf{1.31} & 43.60 & 0.256  & \textbf{1.63} & 43.47 & 0.285  & 0.98 & 43.60 & 0.267  \\
 & PO & \underline{0.98} & 43.34 & 0.291  & 0.98 & 43.73 & \underline{0.295}  & \textbf{1.63} & \underline{43.73} & \underline{0.282}  \\
\rowcolor{egcpshade}  \cellcolor{white} & CLRP & \textbf{1.31} & 37.86 & \underline{0.306}  & \underline{1.31} & 39.03 & 0.226  & \underline{1.31} & 40.21 & 0.259  \\
\bottomrule
\end{tabular}%
}
\caption{Real-Person Retain Set results for LLaVA-1.5-13B across forget ratios and attribute types. Cloze and classification are percentages with two decimal places; generation ROUGE-L keeps three decimals. Higher is better for all metrics. Shaded rows denote CLRP.}
\label{tab:real_retain_image_llava1513b}
\end{table*}

\begin{table*}[t]
\centering
\small
\setlength{\tabcolsep}{3.2pt}
\resizebox{\textwidth}{!}{%
\begin{tabular}{c l ccc ccc ccc}
\toprule
\multirow{2}{*}{\textbf{Ratio}} & \multirow{2}{*}{\textbf{Method}} & \multicolumn{3}{c}{\textbf{Long}} & \multicolumn{3}{c}{\textbf{Numeric}} & \multicolumn{3}{c}{\textbf{Short}} \\
\cmidrule(lr){3-5}\cmidrule(lr){6-8}\cmidrule(lr){9-11}
 & & \ClozeUp & \ClassUp & \RougeUp & \ClozeUp & \ClassUp & \RougeUp & \ClozeUp & \ClassUp & \RougeUp \\
\midrule
\multicolumn{11}{c}{\textbf{Qwen2.5-VL-3B}} \\
\midrule
\multirow{5}{*}{\textbf{5\%}} & GA & \textbf{26.80} & \textbf{73.37} & 0.323  & \textbf{27.12} & \textbf{73.50} & 0.323  & \textbf{26.80} & \textbf{73.63} & 0.323  \\
 & GD & 16.99 & 70.37 & \underline{0.444}  & 16.34 & \underline{70.89} & \textbf{0.459}  & 17.32 & \underline{70.50} & \textbf{0.467}  \\
 & KL & \underline{23.53} & 71.28 & 0.360  & \underline{24.51} & 68.28 & 0.363  & \underline{24.84} & 70.23 & 0.361  \\
 & PO & 13.40 & 70.10 & \textbf{0.455}  & 12.09 & 70.23 & \underline{0.449}  & 18.30 & 70.23 & \underline{0.450}  \\
\rowcolor{egcpshade}  \cellcolor{white} & CLRP & 20.26 & \underline{71.41} & 0.369  & 20.26 & 63.05 & 0.395  & 20.26 & 70.37 & 0.346  \\
\midrule
\multirow{5}{*}{\textbf{10\%}} & GA & \textbf{26.80} & \textbf{73.50} & 0.320 & \textbf{26.47} & \textbf{73.50} & 0.319 & \textbf{26.80} & \textbf{73.50} & 0.320 \\
 & GD & 16.01 & 64.49 & \underline{0.431} & 16.67 & 71.54 & \underline{0.446} & 16.67 & 68.93 & \textbf{0.456} \\
 & KL & \underline{25.49} & \underline{73.37} & 0.372 & \underline{25.16} & \underline{72.85} & 0.356 & \underline{24.51} & \underline{73.11} & 0.358 \\
 & PO & 16.67 & 68.93 & \textbf{0.455} & 15.69 & 68.41 & \textbf{0.453} & 15.03 & 67.49 & \underline{0.441} \\
\rowcolor{egcpshade}  \cellcolor{white} & CLRP & 20.26 & 62.66 & 0.382 & 20.26 & 69.06 & 0.388 & 20.26 & 72.32 & 0.432 \\
\midrule
\multirow{5}{*}{\textbf{15\%}} & GA & \textbf{26.47} & \textbf{73.63} & 0.323 & \textbf{27.12} & \textbf{73.50} & 0.322 & \textbf{26.80} & \textbf{73.63} & 0.325 \\
 & GD & 19.61 & 70.89 & \underline{0.442} & 15.03 & 67.89 & \textbf{0.470} & 14.71 & \underline{71.93} & \textbf{0.477} \\
 & KL & \underline{23.53} & 70.23 & 0.366 & \underline{24.51} & \underline{71.41} & 0.362 & \underline{23.20} & 71.54 & 0.364 \\
 & PO & 16.99 & \underline{71.67} & \textbf{0.461} & 16.67 & 67.62 & \underline{0.448} & 15.69 & 69.19 & \underline{0.452} \\
\rowcolor{egcpshade}  \cellcolor{white} & CLRP & 20.26 & 59.27 & 0.383 & 20.26 & 65.80 & 0.400 & 20.26 & 61.75 & 0.386 \\
\bottomrule
\end{tabular}%
}
\caption{Real-Person Retain Set results for Qwen2.5-VL-3B across forget ratios and attribute types. Cloze and classification are percentages with two decimal places; generation ROUGE-L keeps three decimals. Higher is better for all metrics. Shaded rows denote CLRP.}
\label{tab:real_retain_image_qwen25vl3b}
\end{table*}

\end{document}